%% file: main.tex
\documentclass[final]{l4dc2026}

\input{_macros.tex}

\input{_mathematics.tex}

\input{_definitions.tex}

\title[Belief Net]{Differentiable Filtering for Learning Hidden Markov Models}

\author{%
  \Name{Reginald Zhiyan Chen} \Email{rzchen2@illinois.edu}%
  \AND
  \Name{Heng-Sheng Chang}\thanks{Corresponding author. Coordinated Science Laboratory. Department of Mechanical Science and Engineering.} \Email{hschang@illinois.edu}%
  \AND
  \Name{Prashant G. Mehta} \Email{mehtapg@illinois.edu}\\
  \addr University of Illinois Urbana-Champaign%
}

\begin{document}

  \maketitle

  \begin{abstract}%
    Hidden Markov Models (HMMs) are fundamental for modeling sequential data, yet learning their parameters from observations remains challenging.
    Classical methods like the Baum-Welch algorithm are computationally intensive and prone to local optima, while modern spectral algorithms offer provable guarantees but may produce probability outputs outside valid ranges.
    This work introduces Belief Net, a differentiable filtering framework that learns HMM parameters by formulating the forward filter as a structured neural network and optimizing it with stochastic gradient descent.
    This architecture recursively updates the belief state, which represents the posterior probability distribution over hidden states based on the observation history.
    Unlike black-box transformer models, Belief Net's learnable weights are explicitly the logits of the initial distribution, transition matrix, and emission matrix, ensuring full interpretability.
    The model processes observation sequences using a decoder-only (causal) architecture and is trained end-to-end with standard autoregressive next-observation prediction loss.
    On synthetic HMM data, Belief Net achieves faster convergence than Baum-Welch while successfully recovering parameters in both undercomplete and overcomplete settings, whereas spectral methods prove ineffective in the latter.
    Comparisons with transformer-based models are also presented on real-world language data.
  \end{abstract}


  \section{Introduction} 
  Hidden Markov Models (HMMs) constitute a fundamental class of probabilistic models for discrete-time sequential data, with broad applications spanning speech recognition \citep{rabiner2002tutorial}, natural language processing \citep{manning1999foundations}, computational biology \citep{durbin1998biological}, and financial time series analysis \citep{hassan2005stock}. 
  In an HMM, an observed sequence is generated (emitted) from an unobserved sequence of discrete latent states that evolve as a Markov chain. 
  A time-homogeneous model is fully characterized by three sets of parameters: an initial state distribution, a state transition matrix for the Markov chain, and an emission matrix defining the conditional distribution of observations given latent states \citep{murphy2012machine, elliott1995hidden}.

  The learning problem, or the system identification problem, is to recover the model parameters of the HMM from the observed sequences. 
  A classical approach is the Baum-Welch algorithm \citep{baum1970maximization}, which is a special case of the Expectation-Maximization (EM) algorithm. 
  While widely used, EM is an iterative, non-convex optimization method that is sensitive to initialization and often converges to poor local optima \citep{wu1983convergence}.
  More recently, spectral algorithms have emerged as an efficient and provably correct alternative \citep{hsu2012spectral, boots2011closing, balle2014spectral}.
  These spectral approaches apply singular value decomposition (SVD) to empirical probabilities, the \emph{moments} of observation singles, pairs, and triples, to identify an \emph{observable representation}, yet they often fail in overcomplete regimes due to rank deficiencies and produce outputs that fall outside valid probability ranges \citep{balle2017spectral}.

  In parallel, the field of deep learning has produced powerful general-purpose sequence-to-sequence models, such as transformers \citep{vaswani2017attention}, that excel at next-step prediction through gradient-based optimization \citep{bottou2010large}.
  These models have demonstrated remarkable capabilities in modeling sequential data through an attention mechanism, which captures long-range dependencies and complex patterns in sequences \citep{tay2020long}.
  Unlike an HMM, the parameters of transformer models are not readily interpretable \citep{rudin2019stop}. 
  While they do not explicitly recover the underlying generative structure of the model \citep{lipton2018mythos}, they have consistently achieved superior predictive performance on various tasks \citep{brown2020language, dosovitskiy2020image}.

  Motivated by the relationship between the HMM learning problem and the modeling capabilities of transformer architectures, this paper explores two interrelated questions:
  \begin{itemize}
    \item How well does a transformer perform when the data is generated by an HMM?
    \item How well does an HMM-based learning algorithm perform on real-world language data where transformers excel? 
  \end{itemize}
  To help answer these questions, we introduce a transformer-inspired gradient-based algorithm, referred to as the \emph{Belief Net}, for learning an HMM.
  Closely mirroring the decoder-only architecture of modern auto-regressive language models, Belief Net is designed to perform one-step-ahead prediction by maintaining a \emph{belief state} that encodes the observation history.
  This design choice enables end-to-end training with the same cross-entropy loss used in language modeling, while ensuring that the learned parameters remain interpretable as HMM transition and emission matrices.
  Our contributions are as follows:
  \begin{itemize}
    \item We formulate the HMM's recursive belief state update (the ``forward filter'') as a structured neural network whose learnable weights includes the logits of the initial distribution, transition matrix, and emission matrix.
    \item We show that this model can be trained end-to-end using backpropagation on the standard auto-regressive (next-observation prediction) cross entropy loss function, exactly like a modern decoder-only language model.  
    \item On synthetic data generated by an HMM, we empirically demonstrate that Belief Net is faster than Baum-Welch and can recover parameters in settings where spectral algorithms fail.
    \item On real-world textual data, Belief Net learns an interpretable HMM that serves as a baseline predictive performance against a black-box transformer.  
  \end{itemize}

  The remainder of this paper is organized as follows. 
  Section~\ref{sec:preliminaries} provides a brief overview of HMMs and reviews relevant learning algorithms. 
  Section~\ref{sec:belief_net} introduces the proposed Belief Net framework, including its model architecture and gradient-based parameter optimization scheme. 
  Section~\ref{sec:experiments} presents the experimental evaluation, comparing Belief Net with both classical and modern baselines on synthetic benchmarks and a real-world language modeling task. 
  Section~\ref{sec:conclusion} concludes the paper with a summary of key results and directions for future research.

  \section{Preliminaries and Related Work}\label{sec:preliminaries}

  \subsection{Hidden Markov Models}
  A Hidden Markov Model characterizes a discrete-time stochastic process $\Set{(\stateProcess_t, \observationProcess_t)\in\StateSpace\times\ObservationSpace}_{t\geq0}$, where $\StateSpace = \Set{\state_1, \dots, \state_\stateDimension}$ and $\ObservationSpace = \Set{\observation_1, \dots, \observation_\observationDimension}$ are the hidden (latent) state and observation spaces, respectively.
  For discrete-time steps $t\geq0$, the latent state $\stateProcess_t$ evolves according to the Markov property, while the observation $\observationProcess_t$ is generated conditionally on the current hidden state $\stateProcess_t$.
  The model is fully characterized by the tuple $(\initialDistribution, \transitionMatrix, \emissionMatrix)$:
  The distribution of initial state $\stateProcess_0$ is given by $\initialDistribution(\state)\defined\Probability{\stateProcess_0 = \state}$ for $\state \in \StateSpace$.
  The transition and emission probability matrices are $\transitionMatrix_{ij} \defined \Probability{\stateProcess_{t+1} = \state_j \given \stateProcess_t = \state_i}$ and $\emissionMatrix_{ik} \defined \Probability{\observationProcess_t = \observation_k \given \stateProcess_t = \state_i}$ for $\state_i, \state_j \in \StateSpace$, $\observation_k \in \ObservationSpace$, and $t\geq0$.

  \paragraph{Learning problem} 
  Given a dataset of $N$ observation sequences $\dataset = \inlineset{\observationProcess^{(n)}_{0:\horizon}}_{n=1}^N$ generated by the HMM, whose parameters are unknown to the learner, the goal is to estimate the HMM parameters $(\initialDistribution,\transitionMatrix, \emissionMatrix)$. 
  The number of learnable parameters is $\stateDimension-1$ for $\initialDistribution$, $\stateDimension(\stateDimension-1)$ for $\transitionMatrix$, and $\stateDimension(\observationDimension-1)$ for $\emissionMatrix$, where ($-1$) is because of the normalization constraints on probabilities.
  After being learned \emph{offline}, the HMM can then be applied to a range of inference tasks, including filtering, smoothing, and predicting future observations.

  \subsection{Methods for HMM Learning}
  Several methods have been proposed for learning HMM parameters from observation data:

  \paragraph{Baum-Welch Algorithm} 
  The Baum-Welch algorithm is a classic Expectation-Maximization algorithm used to estimate HMM parameters \citep{baum1970maximization}. 
  It iteratively performs an E-step, which uses the smoothing algorithm to compute the expected state distributions given the observations, followed by an M-step, which re-estimates the parameters ($\initialDistribution, \transitionMatrix, \emissionMatrix$) by calculating the expected state and observation occupancies based on those distributions. 
  The algorithm's objective is to maximize the log-likelihood, and while it is guaranteed to find a local maximum, its performance is sensitive to initialization and it may converge to a poor local optimum.

  \paragraph{Spectral Algorithm} 
  A spectral based algorithm \citep{hsu2012spectral} is employed to identify an observable representation via SVD of empirical probability matrices.
  This approach circumvents local optima and is computationally efficient, as it does not require iterative training. 
  However, its theoretical guarantees rely on a rank condition of the underlying HMM parameters, and the outputs are not necessarily valid probability distributions.
  A simplified version is provided in Appendix~\ref{appx:spectral_algorithm}.

  \paragraph{General Sequence Models (Transformers/RNNs)} 
  Models such as RNNs \citep{rumelhart1985learning}, LSTMs \citep{hochreiter1997long}, and transformers \citep{vaswani2017attention} can be trained on the same next-observation prediction and are highly expressive; 
  however, their parameters do not correspond to the underlying HMM parameters, rendering them uninterpretable black-box models. 
  Related work on learning state-space models with differentiable filtering is reviewed in Appendix~\ref{appx:related_work}.

  \medskip
  A key motivation is to adapt the transformer-based loss function, input-output architecture, and training algorithms, now for learning parameters of an HMM. 
  To help relate the two, the state dimension $\stateDimension$ is set equal to the embedding dimension of a transformer. 
  This is primarily because the number of parameters, e.g., in the attention projection matrices and dense feed-forward networks, scales quadratically with the embedding dimension \citep{geva2021transformer}, analogous to the scaling of parameters in an HMM with respect to the state dimension.

  This design allows for a direct comparison between transformers and Belief Net performance.  
  Such a comparison is useful for two reasons:
  \begin{itemize}
    \item On synthetic data generated by an HMM, it reveals how well a transformer can learn the underlying structure when the data is truly generated by an HMM.
    \item On real-world language data, it quantifies the performance gain that a transformer achieves compared to an interpretable HMM-based algorithm, thereby providing insights into the non-Markovian nature of the embedded latent process in language models.
  \end{itemize}

  \section{The Belief Net Framework}\label{sec:belief_net}
  This paper proposes to learn the HMM parameters $\weights=(\initialDistribution, \transitionMatrix, \emissionMatrix)$ directly by formulating the HMM filter as a transformer-inspired neural network and using stochastic gradient descent to optimize parameters with respect to the standard next-observation prediction loss function, and then compute the estimation (the one-step prediction) $\predictedObservationDistribution_{t+1}(\observation) \defined \Probability{\observationProcess_{t+1}=\observation \given \observationProcess_{0:t}}$ during inference.
  The core idea is to represent the recursive filtering equations as a computational graph \citep{scarselli2008graph, hamilton2022graph}, where the learnable weights correspond directly to the logits of the HMM parameters.
  A comparison with the Baum-Welch algorithm is also presented at the end of this section.

  \begin{figure}[t]
    \vspace{-2\baselineskip}
    \centering
    \begin{tikzpicture}[scale=0.9, every node/.style={scale=0.8}, font=\footnotesize]
      \input{figures/belief_net.tex}
    \end{tikzpicture}
    \caption{
      Belief Net architecture. 
      The model initializes at $t=0$ with $\initialDistribution$ as prior and maintains a belief state $\belief_t$ over the sequence $t\in\Set{0,1,\dots,\horizon-1}$ by recursively applying transition (using transition matrix $\transitionMatrix$) and correction (using previous prior $\belief_{t|t-1}$ and $\emissionProcess_t$ from emission step based on observation $\observationProcess_t$ and emission matrix $\emissionMatrix$) steps to update beliefs.
      The estimation step is to predict probabilities of the next observation $\predictedObservationDistribution_{t+1}$ based on the next prior $\belief_{t+1|t}$ and emission matrix $\emissionMatrix$.
      The detailed computation of each step is described in \algref{alg:belief_net}.
    }\label{fig:belief_net_architecture}
    \vspace{-0.25\baselineskip}
  \end{figure}

  \subsection{Model Architecture}

  \paragraph{Parameterization} Belief Net is parameterized by learnable weights $\tilde{\weights}=(\tilde{\initialDistribution},\tilde{\transitionMatrix},\tilde{\emissionMatrix})$, where $\tilde{\initialDistribution} \in \reals^\stateDimension$ is a row vector of logits for the initial state distribution; 
  $\tilde{\transitionMatrix} \in \reals^{\stateDimension \times \stateDimension}$ is the logits for the transition matrix;
  and $\tilde{\emissionMatrix} \in \reals^{\stateDimension \times \observationDimension}$ is the logits for the emission matrix.
  The softmax operation is applied to obtain valid probability distributions, including 
  \begin{equation}\label{eq:softmax_parameters}
    \initialDistribution = \softmax{\tilde{\initialDistribution}},\quad\transitionMatrix_{i,:} = \inlinesoftmax{\tilde{\transitionMatrix}_{i,:}},\quad\emissionMatrix_{i,:} = \inlinesoftmax{\tilde{\emissionMatrix}_{i,:}},\quad \forall i \in \Set{1,2,\dots,\stateDimension}
  \end{equation}
  The complete set of probability parameters is denoted as $\weights = (\initialDistribution, \transitionMatrix, \emissionMatrix)$, which are functions of the logits $\tilde{\weights}$.
  Thus, for notational simplicity, the transformation from logits to probabilities is denoted as $\weights = \inlinesoftmax{\tilde{\weights}}$.

  \paragraph{HMM Filter}
  For an HMM, the \emph{belief state}, $\belief_t(\state) \defined \Probability{\stateProcess_t = \state \given \observationProcess_{0:t}}$ for $\state \in \StateSpace$, is the posterior distribution over the hidden state, given the history of observations $\observationProcess_{0:t}$, and it is a sufficient statistic for estimating the probability of next observation $\predictedObservationDistribution_{t+1}$.
  The posterior $\belief_t$ is computed recursively using the HMM filter \citep{elliott1995hidden}: 
  for each step $t$, the prior, $\belief_{t|t-1}(\state) \defined \Probability{\stateProcess_t = \state \given \observationProcess_{0:t-1}}$ for $\state \in \StateSpace$, is from the previous step $t-1$.
  It is the distribution over the hidden state $\stateProcess_t$ before observing the current observation $\observationProcess_t$.
  The likelihood is the probability of the current observation given the hidden state, computed through the emission process $\emissionProcess_{t}(\state)\defined\Probability{\observationProcess_{t} \given \stateProcess_t = \state}$ for $\state \in \StateSpace$:
  \begin{itemize}
    \item \emph{Emission Step}: $\emissionProcess_{t}(\state_i) = \emissionMatrix_{i,k}$ observing $\observationProcess_{t} = \observation_k$
  \end{itemize}
  The prior $\belief_{t|t-1}(\state)$ for $\state \in \StateSpace$ is updated using the HMM filter, which consists of two steps:
  \begin{itemize}
    \item \emph{Correction Step:} $\belief_{t}(\state) \propto \emissionProcess_{t}(\state) \belief_{t|t-1}(\state)$
    \item \emph{Transition Step:} $\belief_{t+1|t}(\state) = (\belief_t\transitionMatrix)(\state)$
  \end{itemize}
  The probability of the next observation $\predictedObservationDistribution_{t+1}(\observation)$ for $\observation \in \ObservationSpace$ is then estimated as
  \begin{itemize}
    \item \emph{Estimation Step:} $\predictedObservationDistribution_{t+1}(\observation) = (\belief_{t+1|t}\emissionMatrix)(\observation)$
  \end{itemize}
  This recursive update of the belief state $\belief_t$ forms the core of our proposed model, the \emph{Belief Net}.

  \begin{figure}[t]
    \begin{minipage}[t]{0.47\textwidth}
      \begin{algorithm}[H] 
        \footnotesize
        \caption{Belief Net $\model_{\tilde{\weights}}$}\label{alg:belief_net}
        \setcounter{AlgoLine}{0}
        \input{algorithms/belief_net.tex}
      \end{algorithm}
    \end{minipage}
    \hfill 
    \begin{minipage}[t]{0.49\textwidth}
      \begin{algorithm}[H] 
        \footnotesize
        \caption{Belief Net Learning Process}\label{alg:belief_net_learning}
        \setcounter{AlgoLine}{0}
        \input{algorithms/learning_process.tex}  
      \end{algorithm}
    \end{minipage}
  \end{figure}

  \paragraph{Belief Net Model}
  Given an input sequence $\observationProcess_{0:\horizon-1}$, the model $\model_{\tilde{\weights}}$ maintains belief states $\belief_t$ internally and processes the input observations sequentially using the HMM filter.
  The model then outputs the predicted observation distribution $\predictedObservationDistribution_{t+1}$ for each step $t\in[0,\horizon-1]$.
  Therefore, the model is expressed as 
  \begin{equation*}
    \model_{\tilde{\weights}}:\ObservationSpace^\horizon\to(\ProbabilitySpace{\ObservationSpace})^\horizon,\quad\observationProcess_{0:\horizon-1}\mapsto\model_{\tilde{\weights}}(\observationProcess_{0:\horizon-1})=\predictedObservationDistribution_{1:\horizon}
  \end{equation*}
  where $\ProbabilitySpace{\ObservationSpace}$ denotes the probability simplex over the observation space $\ObservationSpace$.
  The complete model is presented in \algref{alg:belief_net}, and the architecture of the same is illustrated in~\figref{fig:belief_net_architecture}.

  \subsection{Learning Process}
  The Belief Net model $\model_{\tilde{\weights}}$ is trained by minimizing the average cross entropy over all sequences in each mini-batch $\dataset_l\subset\dataset$, a random subset of the full dataset, for each iteration $l$.
  This is identical to the training objective for a decoder-only language model:
  \begin{subequations}
    \label{eq:belief_net_loss}
    \begin{align}
      \loss(\predictedObservationDistribution_{1:\horizon},\observationProcess_{1:T}) &\defined - \frac{1}{\horizon} \sum_{t=1}^{\horizon} \log \predictedObservationDistribution_t(\observationProcess_t)\\
      \mathsf{J}(\tilde{\weights};\dataset_l) &\defined \frac{1}{|\dataset_l|}\sum_{n\in\dataset_l}\loss(\model_{\tilde{\weights}}(\observationProcess^{(n)}_{0:\horizon-1}),\observationProcess^{(n)}_{1:T})
    \end{align}
  \end{subequations}
  The optimization problem is to find the optimal logits $\tilde{\weights}^*$ that minimize the expected loss over the entire dataset:
  \begin{equation}
     \tilde{\weights}^* = \argmin_{\tilde{\weights}} \Expectation[\dataset_l\sim\dataset]{\mathsf{J}(\tilde{\weights};\dataset_l)}
     \label{eq:belief_net_optimization}
  \end{equation}
  This is optimized directly using stochastic gradient descent.
  The estimated model parameters are recovered from the learned logits using the softmax transformation $\hat{\weights} = \inlinesoftmax{\tilde{\weights}^*}$ as in \eqref{eq:softmax_parameters}.
  The complete learning framework is presented in \algref{alg:belief_net_learning}.

  This framework is analogous to the Baum-Welch algorithm, but instead of alternating between E-step (computing expectations) and M-step (maximizing log likelihood), a gradient-based updates is performed on all parameters simultaneously using modern automatic differentiation.
  The detailed comparison between the two algorithms is provided in the next subsection.
  
  \subsection{Comparison with Baum-Welch Algorithm}
  Belief Net and the Baum-Welch algorithm are both iterative approaches for estimating HMM parameters $\weights$ from observed sequences $\observationProcess_{0:\horizon}$. 
  Although they aim to optimize the same objective function, their computation and update mechanisms differ from each other.
  The underlying optimization strategies and computational complexity of the two are discussed in this subsection.

  \paragraph{Objective Function}
  Both algorithms optimize the same objective: the log-likelihood of the observed data with respect to the HMM parameters $\weights$.
  By the chain rule of probability, the log-likelihood is decomposed as
  \begin{equation*}
    \log \Probability{\observationProcess_{0:\horizon} \given \weights} = \sum_{t=0}^\horizon \log \Probability{\observationProcess_t \given \observationProcess_{0:t-1}, \weights}
  \end{equation*}
  where $\observationProcess_{0:-1}$ is defined as the empty sequence.
  This shows that maximizing the joint log-likelihood is equivalent to maximizing the sum of one-step-ahead conditional log-likelihoods.
  Both Baum-Welch and Belief Net aim to find parameters $\weights$ that optimize this objective over the dataset $\dataset$:
  \begin{equation*}
    \max_{\weights} \Expectation[\observationProcess_{0:T}\sim\dataset]{\log \Probability{\observationProcess_{0:\horizon} \given \weights}}
  \end{equation*}
  which is equivalent to minimizing the negative log-likelihood of the observed data with respect to the HMM parameters $\weights$ as in \eqref{eq:belief_net_optimization}.
  The key difference between the two algorithms lies not in the objective function itself, but in how the algorithms optimize it.

  \paragraph{Update Mechanism} The main difference is the strategies for optimizing the same objective.

  \begin{itemize}
    \item \emph{Baum-Welch (EM):} The Baum-Welch algorithm, based on the Expectation-Maximization framework, iteratively maximizes a lower bound $\mathsf{Q}(\weights; \weights^{(l)})$ on the log-likelihood.
    \begin{equation*}
      \log \Probability{\observationProcess_{0:\horizon}|\weights} \geq \mathsf{Q}(\weights; \weights^{(l)}) \defined \Expectation[\Probability{\stateProcess_{0:T}\given\observationProcess_{0:T},\weights^{(l)}}]{\log \Probability{\stateProcess_{0:T},\observationProcess_{0:T}\given\weights}}
    \end{equation*}
    It computes expected sufficient statistics through smoothing and updates parameters in closed form through the necessary condition for maximization:
    \begin{equation*}
      \weights^{(l+1)} = \argmax_{\weights} \mathsf{Q}(\weights; \weights^{(l)})
    \end{equation*}
    \item \emph{Belief Net (Gradient-Based):} Parameters are updated through the optimizer AdamW \citep{loshchilov2017decoupled} with automatic differentiation on the objective \eqref{eq:belief_net_loss}.
    \begin{equation*}
      \tilde{\weights}^{(l+1)} = \tilde{\weights}^{(l)} - \eta_l \mathsf{AdamW}(\nabla_{\tilde{\weights}} \mathsf{J}(\tilde{\weights}^{(l)};\dataset_l)),\quad l=0,1,\dots
    \end{equation*}
    Here, at each iteration $l$, $\eta_l>0$ is the learning rate, $\mathsf{J}(\tilde{\weights}^{(l)};\dataset_l)$ is the log-likelihood over the mini-batch $\dataset_l$, and $\tilde{\weights}^{(l)}$ are the logits corresponding to the HMM parameters.
  \end{itemize}

  \begin{remark}
    The key properties of the expectation-maximization algorithm, including monotonic improvement of the log-likelihood and consistency of convergence to a stationary point, are well-established in the literature \citep{dempster1977maximum, wu1983convergence, krishnamurthy2016partially}.
  \end{remark}

  \begin{remark}
    While gradient-based optimization methods, such as stochastic gradient descent and its variants (e.g., AdamW), do not enjoy the same guarantees on monotonic improvement and consistency of convergence as in the EM algorithm, they have been known for their scalability and flexibility in handling large datasets and complex models \citep{bottou2018optimization}.
  \end{remark}

  \begin{remark}
    Prior work has applied gradient-based methods to learn HMMs without employing the filtering architecture \citep{bagos2004faster}. 
    More recent neural-HMM hybrids further extend this idea \citep{rimella2025hidden}. 
    While sharing the gradient-based learning philosophy, these approaches differ from Belief Net in objectives, architecture, and interpretability.
    More detailed discussion on this neural-HMM hybrid line of work is provided in Appendix~\ref{appx:related_work}.
  \end{remark}

  \paragraph{Computational Complexity}
  The computational complexity of the two algorithms differs on two fronts: the amount of data processed per iteration and the operations required per sequence.
  \begin{itemize}
    \item \emph{Baum-Welch} processes the entire dataset of $N$ sequences in each iteration, requiring full smoothing paths for all sequences.
    \begin{itemize}
      \item Data processed per iteration: $N$ sequences of length $\horizon$.
      \item Operations per sequence: smoothing paths, expected state occupancy, and expected state transition all require $\Order{\horizon\stateDimension^2}$ operations per sequence.
      \item Parameter updates are computed in closed-form with $\Order{\stateDimension^2 + \stateDimension\observationDimension}$ operations.
    \end{itemize}
    The total complexity per iteration is thus $\Order{N\horizon\stateDimension^2}$.

    \item \emph{Belief Net} processes a mini-batch of $B$ sequences each iteration, allowing stochastic updates.
    \begin{itemize}
      \item Data processed per iteration: $B$ sequences of length $\horizon$, where typically $B \ll N$.
      \item Operations per sequence: Both the filtering path and backpropagation require the same order of operations $\Order{\horizon(\stateDimension^2 + \stateDimension\observationDimension)}$ per sequence.
      \item Parameter updates via gradient descent require $\Order{\stateDimension^2 + \stateDimension\observationDimension}$ operations.
    \end{itemize}
    The total complexity per iteration is thus $\Order{B\horizon(\stateDimension^2 + \stateDimension\observationDimension)}$.
  \end{itemize}
  Since typically $B \ll N$, Belief Net processes substantially fewer sequences per iteration, enabling faster iteration times and better scalability.

  \section{Experiments}\label{sec:experiments}
  In this paper, the Belief Net framework is evaluated on two tasks:
  \begin{itemize}
    \item \emph{Synthetic HMM Data:} The objective is to evaluate prediction accuracy and parameter recovery on synthetic data generated from HMMs.
    \item \emph{Real-World Text Data:} The objective is to evaluate prediction performance on text data.
  \end{itemize}
  For both tasks, Belief Net is benchmarked against the Baum-Welch algorithm (from \verb|hmmlearn| library), an independently implemented spectral algorithm, and two transformer-based models: a single-head single-layer model (nanoGPT-s) and a multi-head multi-layer model (nanoGPT-m), both trained from scratch.

  The core objectives are to assess Belief Net's performance compared against both the classical HMM methods and also the transformer architectures, particularly when all models are matched for embedding dimension $\stateDimension$. 
  Studies with varying $\stateDimension$ are also presented.
  Detailed implementation and training setup for all methods are available in Appendix~\ref{appx:implementation_details}.

  \subsection{Synthetic HMM Data: Prediction Accuracy and Parameter Recovery}\label{sec:synthetic_data}
  For the synthetic data, the experimental setting are as follows:

  \paragraph{Data Generation:}
  Each dataset $\dataset=\inlineset{\observationProcess^{(n)}_{0:\horizon}}_{n=1}^{N}$ consists of $N$ sequences of length $\horizon=256$.
  Observation sequences are generated from HMMs with parameters $(\initialDistribution, \transitionMatrix, \emissionMatrix)$ configured as follows:
  \begin{itemize}
    \item \emph{Initial Distribution $\initialDistribution$:} A uniform distribution over the hidden states.
    \item \emph{Transition Matrix $\transitionMatrix$:} A convex combination of a cyclic permutation matrix $\transitionMatrix^{\text{cyclic}}$ ($\state_1 \mapsto \state_2 \mapsto \dots \mapsto \state_\stateDimension \mapsto \state_1$) and a random stochastic matrix $\transitionMatrix^{\text{random}}$: $ \transitionMatrix = \homotopyParameter \transitionMatrix^{\text{cyclic}} + (1-\homotopyParameter) \transitionMatrix^{\text{random}}$, where the homotopy parameter $\homotopyParameter=0.9$ controls the strength of the cyclic structure.
    This structure encourages state persistence while allowing for some randomness in transitions.
    \item \emph{Emission Matrix $\emissionMatrix$:} A random stochastic matrix.
  \end{itemize}
  Both stochastic matrices ($\transitionMatrix^{\text{random}}$ and $\emissionMatrix$) are generated by sampling each entry of the matrix from a normal distribution, then normalizing each row using a softmax function with temperature. The temperatures are set to $0.1$ for $\transitionMatrix^{\text{random}}$ and $0.01$ for $\emissionMatrix$ to control the sparsity of the distributions.
  The validation dataset is generated from the same HMM parameters, but with a smaller number of sequences (10\% of the training set size) to ensure sufficient validation data for model selection.

  \begin{table}[t]
    \vspace*{-2\baselineskip}
    \footnotesize
    \centering
    \input{tables/performance.tex}%

    \vspace{-0.3em}
    \caption{
      Validation loss, number of parameters, and training time on synthetic data for each method.
    }\label{tab:performance_result}
  \end{table}

  \subsubsection{Prediction Accuracy}\label{sec:prediction_accuracy}
  In this setting, the learner knows both the hidden state dimension $\stateDimension$ and the observation dimension $\observationDimension$.
  Two cases are considered:
  \begin{itemize}
      \item \emph{Undercomplete case ($\stateDimension<\observationDimension$):} 
      Hidden states are fewer than observations (e.g., $\stateDimension=64$, $\observationDimension=128$).
      This is the typical scenario where the observation space is richer than the latent state space to which spectral methods are applicable.
      Number of samples in this case is $N=4,000$.
      \item \emph{Overcomplete case ($\stateDimension>\observationDimension$):} 
      Hidden states are more than observations (e.g., $\stateDimension=64$, $\observationDimension=32$).
      This scenario tests whether methods can identify a higher-dimensional latent structure from a limited observation space \citep{sharan2017learning}.
      Number of samples in this case is $N=1,000$.
  \end{itemize}

  Each method's prediction accuracy (cross-entropy loss on a validation set) and training convergence speed (wall-clock time)\footnote{All training computations were conducted on a laptop with an Intel Core i7-12700H processor (2.3GHz, 6P+8E cores), 32GB 4800MHz DDR5 memory, and no GPU acceleration.} were recorded with results summarized in \tabref{tab:performance_result}. 
  The HMM Filter with true parameters is used to establish an optimal loss baseline; and random guessing is used as a worst-case baseline.
  The Belief Net framework consistently outperformed Baum-Welch algorithm, achieving low loss and faster convergence in both under- and overcomplete settings. 
  The spectral method was only effective in the undercomplete case, failing in the overcomplete scenario due to rank deficiencies.
  The nanoGPT models achieved the lowest loss and successfully captured Markovian dependencies, consistent with \citet{hu2024limitation}.
  However, contrary to their findings, the multi-head multi-layer model (nanoGPT-m) performs similarly to the single-head single-layer model (nanoGPT-s), suggesting that a simple architecture suffices to capture the latent Markovian structure.
  Additional results on emperical experiments on sensitivity analyses to initialization are provided in Appendix~\ref{appx:experiments_synthetic_prediction}.

  \subsubsection{Parameter Recovery}\label{sec:parameter_recovery}
  In this realistic setting, the learner only knows the observation dimension $\observationDimension$.
  The dataset $\dataset$ is generated from an HMM with $\stateDimension=64$ and $\observationDimension=128$.
  To evaluate robustness to model misspecification, each model is trained and validated across a range of candidate state dimensions $\hat{\stateDimension} \in \Set{4, 8, 16, 32, 64, 128, 256}$.
  As shown in \figref{fig:parameter_results}, validation loss is minimized when the candidate state dimension $\hat{\stateDimension}$ matches or exceeds the true $\stateDimension$, confirming its effectiveness for model selection. Across all settings, Belief Net outperforms Baum-Welch and spectral methods, approaching the optimal HMM filter when $\hat{\stateDimension} \geq \stateDimension$.
  Additional analyses of eigenvalue spectra and emission matrix discrepancies are provided in Appendix~\ref{appx:experiments_synthetic_parameter}.
  The nanoGPT models achieve the lowest and nearly identical losses, suggesting that a single-head, single-layer architecture suffices to capture the Markovian structure. 
  Notably, nanoGPT-s at $\hat{\stateDimension}=16$ (9,264 parameters) and Belief Net at $\hat{\stateDimension}=64$ (12,352 parameters) attain comparable performance. 
  Additional comparisons of Belief Net with nanoGPT models on various HMM settings are provided in Appendix~\ref{appx:experiments_synthetic_parameter}.

  \begin{figure}[t]
    \vspace{-2.5\baselineskip}
    \centering
    \begin{minipage}{0.49\textwidth}
      \centering
      \begin{tikzpicture}[font=\footnotesize]
        \input{figures/parameter_recovery.tex}
      \end{tikzpicture}
      \vspace{-0.5em}
      \caption{
        Parameter recovery on synthetic data.
        The validation loss $\mathsf{J}$ is plotted with respect to candidate state dimensions $\hat{\stateDimension}$ for each method.
        The true state dimension is $\stateDimension = 64$ and the gray area indicates the $\hat{\stateDimension} \geq \stateDimension$ regime.
        Curves correspond to models (colors): Baum-Welch (blue), Spectral (green), two nanoGPTs (oranges), and Belief Net (red).
        Dashed lines: random guess (gray) and HMM filter (black) represents worst and best scenarios, respectively.
      }\label{fig:parameter_results}
    \end{minipage}
    \hfill
    \begin{minipage}{0.49\textwidth}
      \centering
      \begin{tikzpicture}[font=\footnotesize]
        \input{figures/language_model/loss.tex}
      \end{tikzpicture}
      \vspace{-0.5em}
      \caption{
        Language modeling results on Federalist Papers. 
        The Belief Net's training (light red) and validation (red) loss $\mathsf{J}_l$ over iterations $l$ are shown in solid curves.
        The validation loss is evaluated every fifty iterations.
        For comparison, horizontal dashed lines show the final validation losses achieved by other methods, including random guess (gray), Baum-Welch (blue), Spectral (green), and two nanoGPTs (oranges).
        Corresponding $\mathsf{Perplexity}$ is shown on the right.
      }\label{fig:text_results}
    \end{minipage}
  \end{figure}
  \vspace{-0.5em}

  \subsection{Real-World Text Data: Character-Level Language Modeling}\label{sec:real_world_text}
  Belief Net is evaluated on a language modeling task for next-token prediction using the Federalist Papers dataset \citep{jeong2025small, bhatia2023watch}.
  After tokenization, the dataset comprises $\observationDimension = 82$ unique characters, with $N = 4,000$ training sequences; each sequence has a length of $\horizon = 256$. 
  The latent state dimension is fixed at $\stateDimension = 64$ across all methods.
  \figref{fig:text_results} depicts the training and validation losses, together with validation $\mathsf{Perplexity} \defined e^{\mathsf{J}}$.
  The nanoGPT-m achieves the lowest perplexity, as expected given its capacity to model non-Markovian language structure. Belief Net outperforms classical methods, indicating improved modeling of Markovian latent dynamics, while retaining interpretability: its learned emission matrix captures the dominance of lowercase letters and identifies distinct states for uppercase and digit emissions, demonstrating recovery of meaningful latent structure. Further details are provided in Appendix~\ref{appx:experiments_text}.

  \section{Conclusion}\label{sec:conclusion}
  This paper introduces Belief Net, a differentiable filtering framework that bridges system identification and neural representation learning.
  Experiments show that it outperforms Baum-Welch in convergence speed, recovering parameters even in overcomplete regimes where spectral methods fail. 
  On real text data, Belief Net achieves better performance compared to classical methods.
  Future work will extend the framework to more general settings, including POMDPs, non-Markovian models with memory beyond a single time step, and online learning, thereby broadening its applicability to a wider class of sequential inference and decision-making problems.

  \pagebreak

  \vspace*{-4\baselineskip}

  \acks{%
    This work is supported in part by the AFOSR award FA9550-23-1-0060 and the NSF award 2336137.
    We gratefully acknowledge Dr. Tixian Wang for his valuable insights into nanoGPT models and for his assistance with the initial numerical experiments.
  }

  \bibliography{references}

  \appendix

  \section*{Appendix}
  The appendix provides additional details on the related work, implementation details, and additional experimental results.
  \begin{itemize}
    \item \emph{Related Work:} A review of related literature on learning state-space models and connections to neural network architectures.
    \item \emph{Implementation:} Details on the implementation of each model, including library usage, training procedure, and hyperparameters settings.
    \item \emph{Experiments:} Additional experiment results complementing those in the main text, including training curves, initialization sensitivity analyses, evaluations of parameter recovery, additional comparisons with Belief Net and nanoGPTs, and interpretations of the learned models.
  \end{itemize}

  \input{appendices/related-work.tex}

  \input{appendices/implementation.tex}

  \input{appendices/experiments.tex}

\end{document}

%% file: _macros.tex
\usepackage{mathtools}
\usepackage{times}
\usepackage{booktabs}
\usepackage{multicol}
\usepackage{multirow}
\usepackage{tikz}
\usepackage{pgfplots}
\pgfplotsset{compat=1.18} 
\usepackage{pgfplotstable}
\usetikzlibrary{positioning, arrows.meta, calc, fadings, decorations.markings,pgfplots.fillbetween}

\tikzfading[name=fadeing,
  left color=transparent!40,
  right color=transparent!0]

\hypersetup{
  colorlinks=true,
  linkcolor=black,
  citecolor=black,
  urlcolor=black,
}

\usepackage{xparse}
\usepackage{ifthen}
\usepackage{pifont}

\newcommand{\xmark}{\ding{55}}%

\usepackage{caption-light}
\usepackage[framemethod=TikZ]{mdframed}

\usepackage{algorithm}
\LinesNumbered

\let\oldeqref\eqref
\renewcommand{\eqref}[1]{equation~\oldeqref{#1}}
\newcommand{\algref}[1]{Algorithm~\ref{#1}}
\newcommand{\figref}[1]{Figure~\ref{#1}}
\newcommand{\tabref}[1]{Table~\ref{#1}}

%% file: _mathematics.tex
\newcommand{\defined}{\vcentcolon=}

\newcommand{\reals}{\mathbb{R}}

\newcommand{\Mat}[1][]{\ifthenelse{\equal{#1}{}}{\textnormal{Mat}}{\textnormal{Mat}(#1)}}

\newcommand{\tangent}[1]{
    \ifthenelse{\equal{#1}{}}
    {{T}}
    {{T_{#1}}}
}
\newcommand{\dualtangent}[1]{
    \ifthenelse{\equal{#1}{}}
    {{T^*}}
    {{T_{#1}^*}}
}

\newcommand{\Adjoint}[1]{
    \ifthenelse{\equal{#1}{}}
    {\textnormal{Ad}}
    {\textnormal{Ad}_{#1}}
}

\newcommand{\adjoint}[1]{
    \ifthenelse{\equal{#1}{}}
    {\textnormal{ad}}
    {\textnormal{ad}_{#1}}
}
\newcommand{\coAdjoint}[1]{
    \ifthenelse{\equal{#1}{}}
    {\textnormal{Ad}^*}
    {\textnormal{Ad}^*_{#1}}
}

\newcommand{\coadjoint}[1]{
    \ifthenelse{\equal{#1}{}}
    {\textnormal{ad}^*}
    {\textnormal{ad}^*_{#1}}
}

\newcommand{\Set}[1]{\left\{#1\right\}}
\newcommand{\inlineset}[1]{\{#1\}}
\newcommand{\given}{\,\middle|\,}

\newcommand{\softmax}[1]{\textnormal{softmax}\left(#1\right)}
\newcommand{\inlinesoftmax}[1]{\textnormal{softmax}(#1)}

\newcommand{\inline}[1]{%
  \begingroup
    \let\left\relax
    \let\right\relax
    #1%
  \endgroup
}

\newcommand{\abs}[1]{\left|#1\right|}
\newcommand{\norm}[1]{\left\lVert#1\right\rVert}

\newcommand{\pseudoinverse}{+}

\newcommand{\transpose}{\intercal}

\newcommand{\diag}[1]{\textnormal{diag}\left\{#1\right\}}

\newcommand{\Matrix}[1]{\mathrm{#1}}
\newcommand{\identity}[1]{
    \ifthenelse{\equal{#1}{}}
    {\Matrix{I}}
    {\Matrix{I}_{#1}}
}

\newcommand{\ProbabilitySpace}[1]{\mathcal{P}\left(#1\right)}

\NewDocumentCommand{\Probability}{o m}{
  \mathsf{P}
  \IfValueT{#1}{_{#1}}
  \IfNoValueT{#1}{}
  \left(#2\right)
}
\NewDocumentCommand{\Expectation}{o m}{
  \mathsf{E}
  \IfValueT{#1}{_{#1}}
  \IfNoValueT{#1}{}
  \left(#2\right)
}
\NewDocumentCommand{\Variance}{o m}{
  \mathsf{Var}
  \IfValueT{#1}{_{#1}}
  \IfNoValueT{#1}{}
  \left(#2\right)
}

\DeclareMathOperator*{\argmax}{arg\,max}
\DeclareMathOperator*{\argmin}{arg\,min}

\newcommand{\subjectto}{\textnormal{s.t.}}

\newcommand{\Order}[1]{O\left(#1\right)}

%% file: _definitions.tex
\newcommand{\state}{\mathsf{x}}
\newcommand{\observation}{\mathsf{z}}
\newcommand{\StateSpace}{\mathbb{S}}
\newcommand{\ObservationSpace}{\mathbb{O}}
\newcommand{\stateDimension}{\mathsf{d}}
\newcommand{\observationDimension}{\mathsf{m}}

\newcommand{\stateProcess}{X}
\newcommand{\observationProcess}{Z}

\newcommand{\distribution}{\mu}
\newcommand{\initialDistribution}{\distribution}
\newcommand{\transitionMatrix}{A}

\newcommand{\emissionMatrix}{C}

\newcommand{\emissionProcess}{e}

\newcommand{\VarianceConditioned}[2]{\mathsf{V}_{#1}\ifthenelse{\equal{#2}{}}{}{\left[#2\right]}}

\newcommand{\belief}{\mu}

\newcommand{\predictedObservationDistribution}{p}

\newcommand{\loss}{\ell}

\newcommand{\horizon}{T}

\newcommand{\dataset}{\mathcal{D}}

\newcommand{\weights}{\theta}
\newcommand{\model}{f}

\newcommand{\homotopyParameter}{\alpha}
\newcommand{\eigenValue}{\lambda}
\newcommand{\discrepancy}{\delta}
\newcommand{\coupling}{\gamma}
\newcommand{\couplingSet}{\Gamma}
\newcommand{\stationaryEmissionDistribution}{\bar{p}}

%% file: figures/belief_net.tex
\tikzset{
  block/.style={rectangle, fill=gray!30, minimum width=2cm, minimum height=0.8cm, rounded corners=3pt},
  layer/.style={rectangle, draw, minimum width=2cm, minimum height=0.8cm, rounded corners=3pt, thick},
  dashed-block/.style={rectangle, draw, dashed, minimum width=2cm, minimum height=0.8cm, rounded corners=3pt, thick},
  arrow/.style={-{Triangle[length=2mm]}, thick, rounded corners=5pt},
  node distance=0.8cm and 1.25cm 
}

\def\boxheight{0.8cm}

\def\vdistance{0.8cm}

\node[block] (emission-t) {Emission};
\node[block, above=of emission-t] (correction-t) {Correction};
\node[block, above=of correction-t] (transition-t) {Transition};
\node[block, above=of transition-t] (estimation-t) {Estimation};

\node (center-point) at ($(correction-t)!0.5!(transition-t)$) {};

\draw[arrow] ($(emission-t.south)+(0,-\vdistance)$) -- (emission-t.south) node[midway,left] {$\observationProcess_t$};
\draw[arrow] (emission-t.north) -- (correction-t.south) node[midway,left] {$\emissionProcess_t$};
\draw[arrow] (correction-t.north) -- (transition-t.south) node[midway,left] {$\belief_t$};
\draw[arrow] (transition-t.north) -- (estimation-t.south) node[midway,right] {$\belief_{t+1|t}$};
\draw[arrow] (estimation-t.north) -- ++(0,\vdistance) node[midway, right] {$\predictedObservationDistribution_{t+1}$};

\node[block, right=of emission-t] (emission-t+1) {Emission};
\node[block, right=of correction-t] (correction-t+1) {Correction};

\draw[arrow] ($(emission-t+1.south)-(0,\vdistance)$) -- (emission-t+1.south) node[midway,left] {$\observationProcess_{t+1}$};
\draw[arrow] (emission-t+1.north) -- (correction-t+1.south) node[midway,left] {$\emissionProcess_{t+1}$};
\draw[arrow] (correction-t+1.north) -- ($(correction-t+1.north)+(0,\vdistance)$) node[midway,left] {$\belief_{t+1}$};

\node[block, left=of transition-t] (transition-t-1) {Transition};
\node[block, left=of estimation-t] (estimation-t-1) {Estimation};

\draw[arrow] ($(transition-t-1.south)-(0,\vdistance)$) -- (transition-t-1.south) node[midway,left] {$\belief_{t-1}$};
\draw[arrow] (transition-t-1.north) -- (estimation-t-1.south) node[midway,right] {$\belief_{t|t-1}$};
\draw[arrow] (estimation-t-1.north) -- ++(0,\vdistance) node[midway, right] {$\predictedObservationDistribution_{t}$};

\def\boxpadding{\vdistance/4*1.1}


\def\hdistance{6.5cm}
\def\boxwidth{0.5cm}
\def\dotdistance{1cm}
\def\arrowlength{0.8cm}

\node (box-0) at ($(center-point)+(-\hdistance,0)$) {};
\node[block, minimum width=\boxwidth*2] (emission-0) at (box-0 |- emission-t) {};
\node[block, minimum width=\boxwidth*2] (correction-0) at (box-0 |- correction-t) {};
\node[block, minimum width=\boxwidth*2] (transition-0) at (box-0 |- transition-t) {};
\node[block, minimum width=\boxwidth*2] (estimation-0) at (box-0 |- estimation-t) {};
\node[right=\dotdistance of correction-0] {$\boldsymbol{\dots}$};

\draw[arrow] ($(emission-0.south)+(0,-\vdistance)$) -- (emission-0.south) node[midway,left] {$\observationProcess_0$};
\draw[arrow] (emission-0.north) -- (correction-0.south) node[midway,left] {$\emissionProcess_0$};
\draw[arrow] (correction-0.north) -- (transition-0.south) node[midway,left] {$\belief_0$};
\draw[arrow] (transition-0.north) -- (estimation-0.south) node[midway,right] {$\belief_{1|0}$};
\draw[arrow] (estimation-0.north) -- ++(0,\vdistance) node[midway, right] {$\predictedObservationDistribution_1$};
\draw[arrow] ($(correction-0.west)-(\arrowlength,0)$) -- (correction-0.west) node[near start,above] {$\initialDistribution$};
\draw[arrow] ($(transition-0.north)+(0,\boxpadding/2)$) -| ($(center-point)+(-\hdistance,0)+(\arrowlength,0)$);

\node (box-T) at ($(center-point)+(\hdistance,0)$) {};
\node[block, minimum width=\boxwidth*2] (emission-T) at (box-T |- emission-t) {};
\node[block, minimum width=\boxwidth*2] (correction-T) at (box-T |- correction-t) {};
\node[block, minimum width=\boxwidth*2] (transition-T) at (box-T |- transition-t) {};
\node[block, minimum width=\boxwidth*2] (estimation-T) at (box-T |- estimation-t) {};
\node[left=\dotdistance of transition-T] {$\boldsymbol{\dots}$};

\draw[arrow] ($(emission-T.south)+(0,-\vdistance)$) -- (emission-T.south) node[midway,left] {$\observationProcess_{\horizon-1}$};
\draw[arrow] (emission-T.north) -- (correction-T.south) node[midway,left] {$\emissionProcess_{\horizon-1}$};
\draw[arrow] (correction-T.north) -- (transition-T.south) node[midway,left] {$\belief_{\horizon-1}$};
\draw[arrow] (transition-T.north) -- (estimation-T.south) node[midway,right] {$\belief_{\horizon|T-1}$};
\draw[arrow] (estimation-T.north) -- ++(0,\vdistance) node[midway, right] {$\predictedObservationDistribution_{\horizon}$};
\draw[arrow] ($(center-point)+(\hdistance,0)-(\arrowlength+0.75*\boxwidth,0)$) |- (correction-T.west);
\draw[arrow] ($(transition-T.north)+(0,\boxpadding/2)$) -- ++(1.5*\arrowlength,0);

\def\opacityvalue{0.4}

\shade[left color=white, right color=white, fill opacity=\opacityvalue] ($(estimation-t-1.east)+(0,\boxheight/2+\vdistance)$) rectangle ($(box-0)+(-\boxwidth-4*\boxpadding,-2.5*\vdistance-2*\boxheight)$);
\shade[left color=white, right color=white, fill opacity=\opacityvalue] ($(estimation-t.east)+(0,\boxheight/2+\vdistance)$) rectangle ($(box-T)+(\boxwidth+4*\boxpadding,-2.5*\vdistance-2*\boxheight)$);

\draw[arrow, path fading=fadeing, fading angle=315] ($(transition-t-1.north)+(0,\boxpadding/2)$) -| ($(transition-t-1.east)!0.5!(correction-t.west)$) |- (correction-t.west);
\draw[arrow, path fading=fadeing, fading angle=135] ($(transition-t.north)+(0,\boxpadding/2)$) -| ($(transition-t.east)!0.5!(correction-t+1.west)$) |- (correction-t+1.west);

\def\boxpaddingRatio{0.88}

\draw[layer] ($(center-point)+(-\hdistance-\boxwidth-4*\boxpadding,\vdistance/2+\boxheight+\boxpadding*3*\boxpaddingRatio)$) rectangle ($(center-point)+(\hdistance+\boxwidth+4*\boxpadding,-\vdistance/2-\boxheight-\boxpadding*3*\boxpaddingRatio)$);

%% file: algorithms/belief_net.tex
\SetKwInput{KwParam}{Parameters}

\KwParam{parameters $\tilde{\weights}=(\tilde{\initialDistribution}, \tilde{\transitionMatrix}, \tilde{\emissionMatrix})$}
\KwIn{observation sequence $\observationProcess_{0:\horizon-1}$, }
\KwOut{predicted observation distributions $\predictedObservationDistribution_{1:\horizon}$}
\BlankLine

$(\initialDistribution, \transitionMatrix, \emissionMatrix) \leftarrow \inlinesoftmax{\tilde{\weights}}$\hfill\tcp{to probability}

$\belief_{0|-1} \leftarrow \initialDistribution$\hfill\tcp{prior initialization}
\BlankLine

\For{$t = 0$ \KwTo $\horizon-1$}{
  \tcp{Obtain observation $\observationProcess_t=\observation_k$}
  \BlankLine

  \tcp{Emission (likelihood)}
  $\emissionProcess_t \leftarrow \emissionMatrix_{:,k}$
  \BlankLine

  \tcp{Correction (posterior)}
  $\tilde{\belief}_t \leftarrow \emissionProcess_t \belief_{t|t-1}$

  $\belief_t \leftarrow \tilde{\belief}_t / \text{sum}(\tilde{\belief}_t)$
  \BlankLine

  \tcp{Transition (next prior)}
  $\belief_{t+1|t} \leftarrow \belief_{t}\transitionMatrix$
  \BlankLine
  
  \tcp{Estimation (next observation)}
  $\predictedObservationDistribution_{t+1} \leftarrow \belief_{t+1|t}\emissionMatrix$
}

\Return{$\predictedObservationDistribution_{1:\horizon}$}

%% file: algorithms/learning_process.tex
\KwIn{Dataset $\dataset = \inlineset{\observationProcess^{(n)}_{0:\horizon}}_{n=1}^N$ of observation sequences}
\KwOut{Estimated HMM parameters $\hat{\weights} = (\initialDistribution, \transitionMatrix, \emissionMatrix)$}

\BlankLine

$l \leftarrow 0$\hfill\tcp{iteration counter initialization}
\BlankLine

Initialize learnable parameters $\tilde{\weights}^{(l)}$ randomly

Initialize optimizer $\mathsf{AdamW}$ with learning rate schedule $\eta_l$

\BlankLine

\While{not converged}{
  $\dataset_i \subset \dataset$\hfill\tcp{Sample a mini-batch}
  \BlankLine

  \tcc{Forward: use belief net to compute loss}
  $\mathsf{J}_l \leftarrow \mathsf{J}(\tilde{\weights}^{(l)};\dataset_l)$\hfill\tcp{\eqref{eq:belief_net_loss}}
  \BlankLine

  \tcc{Backward: update parameters through backpropagation}
  $\tilde{\weights}^{(l+1)} \leftarrow \tilde{\weights}^{(l)} - \eta_l \mathsf{AdamW}(\nabla_{\tilde{\weights}} \mathsf{J}_l)$
  \BlankLine

  \tcp{Iteration counter increment}
  $l \leftarrow l + 1$
}

\KwRet $\hat{\weights}=\inlinesoftmax{\tilde{\weights}^{(l)}}$

%% file: tables/performance.tex
\begin{tabular}{cccccccc}
  \toprule
  validation loss & Baum-Welch & Spectral & nanoGPT-s & nanoGPT-m & \textbf{Belief Net} & HMM Filter & Random\\
  \cmidrule(lr){1-1}\cmidrule(lr){2-6}\cmidrule(lr){7-8}
  undercomplete &   1.951   &   1.624  & 1.458 &   1.475   & \textbf{1.569} & 1.368 & 4.852 \\
  overcomplete  &   1.216   &   \xmark  & 0.829 &   0.810   & \textbf{0.830} & 0.737 & 3.466 \\
  \bottomrule
\end{tabular}

\def\horizontalShift{3.47cm}

\begin{tabular}{cccccc}
  \toprule
  \# parameters & Baum-Welch & Spectral & nanoGPT-s & nanoGPT-m & \textbf{Belief Net} \\
  \cmidrule(lr){1-1}\cmidrule(lr){2-6}
  undercomplete & 12,352 & 524,416 & 73,920 & 221,760 &  \textbf{12,352} \\
  overcomplete  & ~~6,208 & \xmark & 51,392 & 199,232 & ~~\textbf{6,208} \\
  \bottomrule
\end{tabular}\hspace*{\horizontalShift}

\begin{tabular}{cccccc}
  \toprule
  training time & Baum-Welch & Spectral & nanoGPT-s & nanoGPT-m & \textbf{Belief Net} \\
  \# iterations & 20 iter. & (PCA) & 2,000 iter. & 2,000 iter. & \textbf{2,000 iter.} \\
  \cmidrule(lr){1-1}\cmidrule(lr){2-6}
  undercomplete & 51 min & N/A & 2.5 min & 7.5 min &  \textbf{20 min} \\
  overcomplete  & 12 min & \xmark & ~~~1 min & 1.5 min & ~~\textbf{6 min} \\
  \bottomrule
\end{tabular}\hspace*{\horizontalShift}

\begin{tikzpicture}[remember picture, overlay]
  \node[anchor=west, xshift=3.3cm, yshift=1.85cm] at (0,0) {
    \begin{minipage}{4cm}
      \footnotesize
      \begin{itemize}
        \item reference loss:
        \begin{itemize}
          \setlength{\itemindent}{-1.6em}
          \item optimal: HMM Filter\\
          \hspace*{-1.6em}(model known)

          \item random: $\ln(\observationDimension)$
        \end{itemize}
        
        \item \xmark: fail\\
        (rank deficiency)
        
        \item  N/A: PCA is instant
      \end{itemize}
    \end{minipage}
  };
\end{tikzpicture}

%% file: figures/parameter_recovery.tex
\pgfmathsetmacro{\lossRandomGuess}{ln(128)}
\def\lossFilter{1.368} 

\def\dataBaumWelsh{
  (4, 3.824)
  (8, 3.517) 
  (16, 2.951) 
  (32, 2.471) 
  (64, 1.951) 
  (128, 1.492) 
  (256, 1.436)
}
\def\dataSpectral{
  (4, 4.569) 
  (8, 3.750) 
  (16, 3.174) 
  (32, 2.512) 
  (64, 1.755)
}
\def\dataNanoGPTsingle{
  (4, 3.237) 
  (8, 2.348) 
  (16, 1.711) 
  (32, 1.529) 
  (64, 1.458) 
  (128, 1.475) 
  (256, 1.435)
}
\def\dataNanoGPTmulti{
  (4, 3.373) 
  (8, 2.405) 
  (16, 1.628) 
  (32, 1.523) 
  (64, 1.475) 
  (128, 1.433) 
  (256, 1.399)
}
\def\dataBeliefNet{
  (4, 3.758) 
  (8, 3.404) 
  (16, 2.937) 
  (32, 2.332) 
  (64, 1.569) 
  (128, 1.384) 
  (256, 1.381)
}

\definecolor{colorBlue}{RGB}{31,119,180}
\definecolor{colorOrange}{RGB}{255,127,14}
\definecolor{colorGreen}{RGB}{44,160,44}
\definecolor{colorRed}{RGB}{214,39,40}

\def\minStateDim{3}
\def\maxStateDim{512}

\begin{axis}[
  width=7.5cm,
  height=5.35cm,
  xlabel={candidate state dimension $\hat{\stateDimension}$ ($\stateDimension=64$)},
  ylabel={validation loss $\mathsf{J}$},
  xlabel style={yshift=5pt},
  xmode=log,
  log basis x=2,
  xmin=\minStateDim, xmax=\maxStateDim,
  ymin=0, ymax=6,
  xtick={4,8,16,32,64,128,256},
  xticklabels={4,8,16,32,64,128,256},
  grid=major,
  grid style={dashed, gray!20},
  legend pos=north east,
  legend style={
    line width=0.3pt,
  },
  legend cell align={left},
  line width=1.5pt,
  mark size=1.5pt,
  tick label style={font=\footnotesize},
  label style={font=\small},
  every axis plot/.append style={thick},
  axis on top,
]

  \fill[gray!15, opacity=0.6] (axis cs:64,0) rectangle (axis cs:\maxStateDim, 6);

  \addplot[
    color=colorBlue, 
    mark=square,
  ] coordinates \dataBaumWelsh;
  \addlegendentry{\color{black!60}Baum-Welch}

  \addplot[
    color=colorGreen, 
    mark=o,
  ] coordinates \dataSpectral;
  \addlegendentry{\color{black!60}Spectral}

  \addplot[
    color=colorOrange, 
    mark=triangle,
  ] coordinates \dataNanoGPTsingle;
  \addlegendentry{\color{black!60}nanoGPT-s}

  \addplot[
    opacity=0.7,
    color=colorOrange, 
    mark=triangle,
  ] coordinates \dataNanoGPTmulti;
  \addlegendentry{\color{black!60}nanoGPT-m}

  \addplot[
    color=colorRed, 
    mark=diamond,
  ] coordinates \dataBeliefNet;
  \addlegendentry{\textbf{Belief Net}}

  \def\xposition{3.3}

  \node[
    anchor=south west, 
    gray,
  ] at (axis cs:\xposition, \lossRandomGuess) {
    random ($\ln\text{128}$)
  };
  \addplot[
    gray, 
    dashed, 
    line width=1pt, 
    forget plot,
  ] coordinates {
    (\minStateDim, \lossRandomGuess) 
    (\maxStateDim, \lossRandomGuess)
  };

  \node[
    anchor=north west, 
    black,
  ] at (axis cs:\xposition, \lossFilter) {
    HMM filter (optimal)
  };
  \addplot[
    black, 
    dashed, 
    line width=1pt, 
    forget plot,
  ] coordinates {
    (\minStateDim, \lossFilter) 
    (\maxStateDim, \lossFilter)
  };

\end{axis}

%% file: figures/language_model/loss.tex
\def\perplexityRandomGuess{82}
\pgfmathsetmacro{\lossRandomGuess}{ln(\perplexityRandomGuess)}

\def\lossBaumWelch{2.286}
\def\lossSpectral{3.020}
\def\lossNanoGPTsingle{1.492}
\def\lossNanoGPTmulti{1.351}
\def\lossBeliefNet{2.002}

\pgfmathsetmacro{\perplexityBaumWelch}{exp(\lossBaumWelch)}
\pgfmathsetmacro{\perplexitySpectral}{exp(\lossSpectral)}
\pgfmathsetmacro{\perplexityNanoGPTsingle}{exp(\lossNanoGPTsingle)}
\pgfmathsetmacro{\perplexityNanoGPTmulti}{exp(\lossNanoGPTmulti)}
\pgfmathsetmacro{\perplexityBeliefNet}{exp(\lossBeliefNet)}

\definecolor{colorBlue}{RGB}{31,119,180}
\definecolor{colorOrange}{RGB}{255,127,14}
\definecolor{colorGreen}{RGB}{44,160,44}
\definecolor{colorRed}{RGB}{214,39,40}

\def\minIteration{0.5}
\def\maxIteration{6000}

\pgfplotsset{
  empty legend/.style={
    draw=none,
    mark=none,
    line width=0pt,
    forget plot
  }
}

\def\axisWidth{6.3cm}
\def\axisHeight{5.35cm}

\begin{axis}[
  width=\axisWidth,
  height=\axisHeight,
  xlabel={iteration $l$},
  ylabel={loss $\mathsf{J}_l$},
  xlabel style={yshift=5pt},
  xmode=log,
  xmin=\minIteration, xmax=\maxIteration,
  ymin=0, ymax=6,
  xtick={1,10, 100, 1000, 10000},
  grid=major,
  grid style={dashed, gray!20},
  legend pos=north east,
  legend style={
    line width=0.3pt,
  },
  legend cell align={left},
  line width=1.5pt,
  tick label style={font=\footnotesize},
  label style={font=\small},
  every axis plot/.append style={thick},
  axis on top,
  axis y line*=left,
]
  \addlegendimage{empty legend, opacity=0}
  \addlegendentry{\hspace*{-2.15em}\textbf{Belief Net}}


  \def\xpositionLeft{0.7}
  \def\xpositionRight{6000}

  \node[
    anchor=north west, 
    colorBlue,
  ] at (axis cs:\xpositionLeft, \lossBaumWelch) {Baum-Welch};
  \addplot[
    colorBlue, 
    dashed, 
    line width=1pt, 
    forget plot,
  ] coordinates {(\minIteration, \lossBaumWelch) (\maxIteration, \lossBaumWelch)};

  \node[
    anchor=south west, 
    colorGreen,
  ] at (axis cs:\xpositionLeft, \lossSpectral) {Spectral};
  \addplot[
    colorGreen, 
    dashed, 
    line width=1pt, 
    forget plot,
  ] coordinates {(\minIteration, \lossSpectral) (\maxIteration, \lossSpectral)};

  \node[
    anchor=north west, 
    colorOrange,
  ] at (axis cs:\xpositionLeft, \lossNanoGPTsingle) {nanoGPT-s};
  \addplot[
    colorOrange, 
    dashed, 
    line width=1pt, 
    forget plot,
  ] coordinates {(\minIteration, \lossNanoGPTsingle) (\maxIteration, \lossNanoGPTsingle)};

  \node[
    opacity=0.7,
    anchor=north east, 
    colorOrange,
  ] at (axis cs:\xpositionRight, \lossNanoGPTmulti) {nanoGPT-m};
  \addplot[
    opacity=0.7,
    colorOrange, 
    dashed, 
    line width=1pt, 
    forget plot,
  ] coordinates {(50, \lossNanoGPTmulti) (\maxIteration, \lossNanoGPTmulti)};

  \node[
    anchor=south west, 
    gray,
  ] at (axis cs:\xpositionLeft, \lossRandomGuess) {random};
  \addplot[
    gray, 
    dashed, 
    line width=1pt, 
    forget plot,
  ] coordinates {(\minIteration, \lossRandomGuess) (\maxIteration, \lossRandomGuess)};

  \addplot[
    color=colorRed,
    opacity=0.6,
    line width=1pt,
  ] table[
    col sep=comma,  
    x=iteration, 
    y=loss,
  ] {figures/language_model/training_loss.csv};
  \addlegendentry{training}

  \addplot[
    color=colorRed,
    line width=1pt,
  ] table[
    col sep=comma,  
    x=iteration, 
    y=loss,
  ] {figures/language_model/validation_loss.csv};
  \addlegendentry{validation}

\end{axis}

\begin{axis}[
  width=\axisWidth,
  height=\axisHeight,
  xmin=\minIteration, xmax=\maxIteration,
  ymin=0, ymax=6,
  ylabel={$\mathsf{Perplexity}$},
  ylabel style={font=\small},
  axis y line*=right,
  axis x line=none,
  ytick={ 
    \lossRandomGuess,
    \lossSpectral,
    \lossNanoGPTmulti,
    \lossBeliefNet
  },
  yticklabels={
    \pgfmathprintnumber[precision=1]{\perplexityRandomGuess},
    \pgfmathprintnumber[precision=1]{\perplexitySpectral},
    \pgfmathprintnumber[precision=1]{\perplexityNanoGPTmulti},
    \pgfmathprintnumber[precision=1]{\perplexityBeliefNet}
  },
  tick label style={font=\footnotesize},
  line width=1.5pt,
]
\end{axis}

%% file: appendices/related-work.tex
\section{Related Work on Learning State-Space Models}\label{appx:related_work}

This work is closely related to a broad class of architectures that learn state-space models by unrolling recursive updates into computational graphs, thereby enabling end-to-end training with gradient-based optimization \citep{scarselli2008graph, hamilton2022graph}. 
Such approaches have been particularly successful for continuous-state estimators, including neural variants of Kalman filters \citep{revach2022kalmannet}, and more recently within the broader paradigm of differentiable filtering \citep{kloss2021train, wu2025dkfnet}. 
Extensions to nonlinear and non-Gaussian systems have also been developed through differentiable particle filters, which integrate sequential Monte Carlo methods into deep learning frameworks for state inference \citep{brady2024regime}. 
In contrast to these continuous-state formulations, where the belief state is typically represented as a mean-covariance pair or a set of weighted particles, Belief Net operates in a discrete setting, in which the belief state is a probability vector over discrete latent states.

A closely related line of work is Hidden Markov Neural Networks (HMNNs) \citep{rimella2025hidden}, which similarly combine HMM structure with neural architectures. However, the objectives differ fundamentally. 
HMNNs treat neural network weights themselves as latent HMM states, enabling continual learning as new data arrive. 
In contrast, Belief Net leverages a neural computation graph purely for offline system identification: its goals are 
(i) to recover the classical HMM parameters from historical observations, and 
(ii) to perform inference using the standard HMM filtering procedure with the learned parameters. 
In this sense, HMNNs extend what the latent states represent, whereas Belief Net focuses on how HMM parameters are estimated.

%% file: appendices/implementation.tex
\section{Implementation Details}\label{appx:implementation_details}

This section provides detailed implementation specifications for all methods evaluated in Section~\ref{sec:experiments}.
Section~\ref{appx:baum_welch_algorithm} outlines the Baum-Welch algorithm implemented using the \verb|hmmlearn| library.
Section~\ref{appx:spectral_algorithm} describes the spectral algorithm adapted from the \verb|spectral-learning| repository, extended with a custom probability prediction function.
Section~\ref{appx:nanoGPT_model} details the transformer-based models implemented using the \verb|nanoGPT| repository, which offers a minimal implementation of the transformer architecture.
Finally, Section~\ref{appx:belief_net} presents the Belief Net implementation using \verb|PyTorch| for automatic differentiation and neural network computations.

\subsection{Baum-Welch Algorithm}\label{appx:baum_welch_algorithm}

The \texttt{hmmlearn} Python library~\citep{hmmlearn2024python} was utilized for all experiments involving the Baum-Welch algorithm.
For each run, a \texttt{CategoricalHMM} object was configured with randomly initialized parameters, including the initial distribution $\initialDistribution$, transition matrix $\transitionMatrix$, and emission matrix $\emissionMatrix$, and a maximum of 20 iterations was specified. 
The model was then trained on the dataset $\dataset$ using the \texttt{fit} method, which iteratively estimates the HMM parameters $\weights = (\initialDistribution, \transitionMatrix, \emissionMatrix)$. 
Upon reaching the iteration limit, the learned parameters $\hat{\weights} = (\hat{\initialDistribution}, \hat{\transitionMatrix}, \hat{\emissionMatrix})$ were extracted and used to evaluate the validation loss.
To reduce sensitivity to random initialization, the procedure was repeated 5 times with distinct random seeds, and the run with the lowest validation loss was selected.

\subsection{Spectral Algorithm}\label{appx:spectral_algorithm}

The spectral algorithm implementation follows the method described in~\citet{hsu2012spectral}. 
The model initialization and parameter estimation were adapted from \verb|spectral-learning| \citep{spectrallearning2014python}, while the probability prediction was based on the original paper \citep{hsu2012spectral}.

\paragraph{Empirical Probabilities} 
Given training data $\dataset$, construct the empirical probability estimates $P_1, P_{2,1}, P_{3,k,1}$ by counting number of occurrences of overlapping subsequences of length three in the training data.
For any observation $\observation,\observation' \in \ObservationSpace$ and for any $t$ within the sequence, the following time-invariant empirical probability estimates are computed:
\begin{itemize}
  \item Probability vector of dimension $\observationDimension$: $P_{1}(\observation) = \Probability{\observationProcess_t=\observation\given\dataset}$
  \item Probability matrix of dimension $\observationDimension \times \observationDimension$: $P_{2,1}(\observation', \observation) = \Probability{\observationProcess_{t+1}=\observation', \observationProcess_t=\observation\given\dataset}$
  \item Matrix-valued probability function of dimension $\observationDimension \times \observationDimension$: 
  \begin{equation*}
    \ObservationSpace\to\ProbabilitySpace{\ObservationSpace^2},\quad\observation_k\mapsto P_{3,k,1}(\observation', \observation) \propto \Probability{\observationProcess_{t+2}=\observation', \observationProcess_{t+1}=\observation_k, \observationProcess_t=\observation\given\dataset}
  \end{equation*}
\end{itemize}

\paragraph{Observable Representation}
To compute the observable representation, the SVD of $P_{2,1}$ is first performed: $P_{2,1} = \Matrix{U}\Sigma\Matrix{V}^\transpose$, where $\Matrix{U}\in \reals^{\observationDimension \times \observationDimension}$ and $\Matrix{V}\in \reals^{\observationDimension \times \observationDimension}$ are orthogonal matrices, and $\Sigma=\diag{\sigma_1,\sigma_2,\dots,\sigma_\observationDimension}$ is a diagonal matrix of singular values $\sigma_k$ sorted in descending order.  
The top $\stateDimension$ singular values $\sigma_{1:\stateDimension}$ are retained, and the corresponding left singular vectors are extracted to form the matrix $\Matrix{U}_{\stateDimension}\in \reals^{\observationDimension \times \stateDimension}$. 
The observable representation parameters are computed as follows:
\begin{equation*}
b_0 = \Matrix{U}_\stateDimension^\transpose P_{1},\quad 
b_\infty = (P_{2,1}^{\transpose}\Matrix{U}_\stateDimension)^\pseudoinverse P_1,\quad
\Matrix{B}_k = \Matrix{U}_\stateDimension^\transpose P_{3,k,1} (\Matrix{U}_\stateDimension^\transpose P_{2,1})^{\pseudoinverse}, \quad \forall\observation_k \in \ObservationSpace
\end{equation*}
where the superscript $\pseudoinverse$ denotes the Moore-Penrose pseudoinverse.

\paragraph{Recursive Update} 
Once obtaining a new observation $\observationProcess_t$, the internal state $b_{t}$ is updated following:
\begin{equation*}
    b_{t+1} = \frac{\Matrix{B}_k b_t}{b_\infty^{\transpose} \Matrix{B}_k b_t},\quad \text{with }\observationProcess_{t} = \observation_k
\end{equation*}
where the case of denominator being zero is handled by resetting $b_{t+1} = b_0$.

\paragraph{Prediction} 
The conditional probability for the next observation $\observationProcess_{t+1}$ given history $\observationProcess_{0:t}$ is:
\begin{equation*}
\Probability{\observationProcess_{t+1}=\observation_k \given \observationProcess_{0:t}} \propto b_\infty^\transpose \Matrix{B}_k b_t,\quad \forall \observation_k \in \ObservationSpace
\end{equation*}
To handle the cases of negative probabilities, all negative value entries were replaced with the minimum positive value at the current step, and the resulting vector was renormalized to sum to unity.

\subsection{transformer Model}\label{appx:nanoGPT_model}

The transformer baselines were implemented using the \verb|nanoGPT| repository~\citep{karpathy2024nanogpt}, which provides a minimal decoder-only GPT architecture.
Two configurations were evaluated to assess the impact of model capacity on performance:
\begin{itemize}
  \item nanoGPT-s: A single-head, single-layer transformer represents the simplest architecture.
  \item nanoGPT-m: A multi-head, multi-layer transformer (4 heads, 4 layers) represents a more expressive variant.
\end{itemize}
Both models were configured with an embedding dimension of $\stateDimension$, a feed-forward dimension of $4\stateDimension$, and learnable positional embeddings. 
Training was performed on a next-observation prediction task using cross-entropy loss. 
The hyperparameters included a batch size of 10, dropout values in $\Set{0.0, 0.1}$, learning rates in $\Set{0.001, 0.01}$, and a maximum of 2,000 iterations for synthetic data and 4,000 iterations for text data. 
A grid search over the dropout and learning rate values was conducted, with the configuration achieving the lowest validation loss selected for final evaluation. 
All remaining settings followed the repository defaults for CPU-only training. 
To support the experimental setup, the \texttt{train.py} script was modified to allow direct data loading from generated sequences, while the core model architecture defined in \texttt{model.py} remained unchanged.

\subsection{Belief Net}\label{appx:belief_net}

The repositories containing the Belief Net implementation and the experiments presented in Section~\ref{sec:experiments} are available in \citep{chang2025signalsystem, chen2026beliefnet} and were developed as part of this work. 
The implementation follows Algorithms~\ref{alg:belief_net} and \ref{alg:belief_net_learning} in the main text. 
The model is trained using a batch size of 10, with dropout and learning rate selected from $\Set{0.0, 0.1}$ and $\Set{0.01, 0.1}$, respectively, via grid search based on validation loss. 
Training is run for up to 2,000 iterations on synthetic data and 4,000 iterations on text data. 
Optimization is performed using AdamW with betas $(0.9, 0.999)$, eps $10^{-8}$, and weight decay $0.01$, without AMSGrad, and with decoupled weight decay enabled.

\begin{remark}
  A theoretical analysis of the convergence and sample complexity of the gradient-based Belief Net is an important open direction.
  Existing results for stochastic gradient descent on smooth non-convex objectives~\citep{ghadimi2013stochastic} guarantee convergence to a stationary point at a rate of $\inline{\Order{1/\sqrt{L}}}$ after $L$ iterations, but translating these into sample-complexity bounds specific to HMM parameter recovery requires additional structural assumptions (e.g. identifiability and mixing conditions) that are beyond the scope of the present work and are deferred to future research.
\end{remark}

%% file: appendices/experiments.tex
\section{Experiments Results}\label{appx:experiments}
This section presents extra results that complement the findings reported in Section~\ref{sec:experiments}.
The results are organized into two main subsections: synthetic HMM data in Section~\ref{appx:experiments_synthetic} and real-world text data in Section~\ref{appx:experiments_text}, corresponding to Section~\ref{sec:synthetic_data} and Section~\ref{sec:real_world_text} in the main text, respectively.
Each subsection includes additional figures and tables that provide a more comprehensive view of the models' performance across different settings and metrics.

\subsection{Synthetic HMM Data: Prediction Accuracy and Parameter Recovery}\label{appx:experiments_synthetic}

\subsubsection{Prediction Accuracy}\label{appx:experiments_synthetic_prediction}

\begin{figure}[t]
  \centering

  \hfill
  \begin{minipage}{0.44\textwidth}
    \centering
    \begin{tikzpicture}[font=\footnotesize]
      \input{figures/hmm/undercomplete/loss.tex}
    \end{tikzpicture}
  \end{minipage}
  \hspace*{0.1em}
  \begin{minipage}{0.45\textwidth}
    \centering
    \begin{tikzpicture}[font=\footnotesize]
      \input{figures/hmm/overcomplete/loss.tex}
    \end{tikzpicture}
  \end{minipage}%
  \hspace*{-0.2em}

  \hfill
  \begin{minipage}[t]{0.88\textwidth}
    \vspace{0.25\baselineskip}
    \centering
    \small
    iteration $l$
  \end{minipage}
  
  \centering
  \begin{tikzpicture}[remember picture, overlay]
    \node[anchor=east, xshift=-4.4cm, yshift=0.6cm] at (0,0) {
      \begin{minipage}{4cm}
        \begin{center}
          \begin{tikzpicture}[font=\footnotesize]

            \definecolor{colorBlue}{RGB}{31,119,180}
            \definecolor{colorOrange}{RGB}{255,127,14}
            \definecolor{colorGreen}{RGB}{44,160,44}
            \definecolor{colorRed}{RGB}{214,39,40}
            \begin{axis}[
                hide axis,
                xmin=10, xmax=50, ymin=0, ymax=0.4,
                legend style={
                    draw=none,
                    fill=none,
                    at={(0.5,0.5)}, 
                    anchor=center, 
                    cells={anchor=west}, 
                    legend columns=1, 
                    /tikz/every even column/.append style={column sep=5pt},
                },
            ]

              \addlegendimage{empty legend, opacity=0}
              \addlegendentry{\hspace*{-2.15em}\textbf{Belief Net}}

              \addlegendimage{
                color=colorRed, 
                opacity=0.6,
                line width=1.5pt,
              }
              \addlegendentry{training}

              \addlegendimage{
                color=colorRed,
                line width=1.5pt,
              }
              \addlegendentry{validation}

            \end{axis}
          \end{tikzpicture}
        \end{center}
      \end{minipage}
    };
  \end{tikzpicture}

  \caption{
    Undercomplete and overcomplete results on synthetic data.
    The Belief Net's training (light red) and validation (red) loss $\mathsf{J}_l$ over iterations $l$ are shown in solid curves.
    The validation loss is evaluated every 50 iterations.
    For comparison, horizontal dashed lines show the final validation losses achieved by other methods, including random guess (gray), Baum-Welch (blue), Spectral (green), and nanoGPT (orange).
    }
  \label{fig:undercomplete_overcomplete_training_results}
  \vspace{0.25\baselineskip}
\end{figure}

\begin{figure}[t]
  \centering
  \input{figures/initialization/initialization_sensitivity.tex}
  \caption{
    Sensitivity to initialization for the Belief Net (red) and Baum-Welch (blue) methods. 
    Both approaches are evaluated in undercomplete and overcomplete regimes across a range of candidate state dimensions $\hat{\stateDimension}$.
    For each setting, 10 independent random initializations are performed, and the resulting validation losses are visualized by their mean (mark), minimum (lower bound), and maximum values (upper bound).
    The horizontal dashed lines indicate the loss of a random guess (gray) and the optimal HMM filter (black), which represent the worst and best scenarios, respectively.
  }\label{fig:initialization_overcomplete}
\end{figure}

In this setting (see Section~\ref{sec:prediction_accuracy}), the learner knows both the hidden state dimension $\stateDimension$ and the observation dimension $\observationDimension$.
The Belief Net's training and validation loss curves for the undercomplete and overcomplete settings on synthetic HMM data are shown in \figref{fig:undercomplete_overcomplete_training_results}.

To evaluate sensitivity to initialization, we relax the assumption that the hidden state dimension $\stateDimension$ is known and instead sweep over candidate state dimensions $\hat{\stateDimension} \in \Set{4, 8, 16, 32, 64, 128, 256}$. 
For each candidate, we assess the variability in validation loss $\mathsf{J}$ across multiple random initializations. 
The results are shown in \figref{fig:initialization_overcomplete}. The Belief Net consistently achieves lower validation loss and exhibits substantially less variability compared to Baum-Welch across all candidate state dimensions. 
The spectral method is excluded from this analysis, as it does not depend on initialization.

\subsubsection{Parameter Recovery}\label{appx:experiments_synthetic_parameter}

\begin{figure}[t]
  \vspace*{-2\baselineskip}

  \centering
  \input{figures/parameter_recovery/matrix_comparison.tex}
  \vspace{0.25\baselineskip}
  \caption{
    Parameter recovery on synthetic HMM data comparing the Belief Net (red) and Baum-Welch (blue) methods. 
    The top row illustrates the magnitudes of the eigenvalues for the learned transition matrices, arranged in descending order across multiple candidate state dimensions ($\hat{\stateDimension} \in \Set{4, 16, 64, 256}$), where transparency levels differentiate the choices of $\hat{\stateDimension}$ and black markers denote the true transition matrix eigenvalues for reference. 
    The bottom row evaluates the fidelity of the estimation by reporting the discrepancy between the learned and ground-truth emission matrices as a function of the candidate state dimension $\hat{\stateDimension}$.
  }
  \label{fig:parameter_recovery_results}
\end{figure}

In this setting (see Section~\ref{sec:parameter_recovery}), the learner only knows the observation dimension $\observationDimension$.
Each model is trained and validated across a range of candidate state dimensions $\hat{\stateDimension} \in \Set{4, 8, 16, 32, 64, 128, 256}$.
To evaluate the quality of parameter recovery, we analyze the learned transition and emission matrices from the Belief Net and Baum-Welch methods, where the results are shown in \figref{fig:parameter_recovery_results}.
For the transition matrix, we compare the magnitudes of the eigenvalues of the learned transition matrices $\inline{\abs{\eigenValue(\hat{\transitionMatrix})}}$ to those of the true transition matrix $\inline{\abs{\eigenValue(\transitionMatrix)}}$, which reflect the system's temporal dynamics and mixing properties.
For $\hat{\stateDimension} > \stateDimension$, both the Belief Net and Baum-Welch successfully recovers the eigenvalue spectrum of the transition matrix, with the learned $\stateDimension$ eigenvalues closely matching the true eigenvalues and the remaining $\hat{\stateDimension}-\stateDimension$ eigenvalues have a sharp drop in magnitude, indicating that the extra dimensions are effectively ignored.
For $\hat{\stateDimension} \leq \stateDimension$, the learned eigenvalues deviate from the true eigenvalues, reflecting the model's inability to capture the full dynamics with insufficient state dimensions.
For the emission matrix, we evaluate the discrepancy $\discrepancy$ between the learned and true emission matrices as a function of the candidate state dimension $\hat{\stateDimension}$.
For a fixed $\hat{\stateDimension}$, the discrepancy is computed as follows:
\begin{equation*}
  \discrepancy(\hat{\emissionMatrix}, \emissionMatrix) = \sum_{k=1}^{\observationDimension}\stationaryEmissionDistribution(\observation_k)\sum_{i=1}^{\hat{\stateDimension}}\sum_{j=1}^{\stateDimension} \coupling_{ij}\abs{\hat{\emissionMatrix}_{ik} - \emissionMatrix_{jk}}
\end{equation*}
where $\stationaryEmissionDistribution\in\ProbabilitySpace{\ObservationSpace}$ is the stationary distribution of the true HMM's emission process, and $\coupling \in\couplingSet\defined [0,1/{\stateDimension}]^{\hat{\stateDimension}} \times [0,1/{\hat{\stateDimension}}]^\stateDimension$ is an optimal coupling between the rows of the learned emission matrix $\hat{\emissionMatrix}$ and the true emission matrix $\emissionMatrix$.
The coupling $\coupling$ is computed by solving the following optimal transport problem:
\begin{equation*}
  \begin{aligned}
    \min_{\coupling\in\couplingSet}~\sum_{i=1}^{\hat{\stateDimension}}\sum_{j=1}^{\stateDimension} \coupling_{ij}\norm{\hat{\emissionMatrix}_{i,:} - \emissionMatrix_{j,:}},\quad\subjectto~\sum_{i=1}^{\hat{\stateDimension}} \coupling_{ij} = \frac{1}{\stateDimension},~\forall j=1,\dots,\stateDimension,\quad\sum_{j=1}^{\stateDimension} \coupling_{ij} = \frac{1}{\hat{\stateDimension}},~\forall i=1,\dots,\hat{\stateDimension}
  \end{aligned}
\end{equation*}
where the norm $\norm{\cdot}$ can be any metric between the rows of the emission matrices.
Here, we use the Hellinger distance, which is a common choice for measuring the distance between probability distributions, and the optimal coupling is obtained using the Sinkhorn algorithm~\citep{sinkhorn1967diagonal} through the Python Optimal Transport (\texttt{POT}) library~\citep{flamary2021pot}.
The results show that the Belief Net almost consistently achieves lower emission matrix discrepancy compared to Baum-Welch across all candidate state dimensions $\hat{\stateDimension}$, indicating better parameter recovery quality.

\begin{figure}[t]
  \centering
  \input{figures/parameter_recovery/complete_comparison.tex}
  \caption{
    Belief Net (red) v.s. nanoGPTs (orange) on synthetic HMM data with state dimension $\stateDimension=64$. 
    This figure compares the models' validation loss $\mathsf{J}$ across candidate state dimensions $\hat{\stateDimension}$ for HMM's with both fast and slow mixing, overcomplete and undercomplete settings. 
    Random guess loss (gray) and optimal loss (black) are plotted for worst and best-case comparison.
  }
  \label{fig:nanoGPTs_HMMs}
\end{figure}

In addition to comparing the Belief Net and Baum–Welch methods, we evaluate the Belief Net against nanoGPT-s and nanoGPT-m across varying candidate state dimensions $\hat{\stateDimension}$ for both fast- and slow-mixing HMMs. 
As shown in \figref{fig:nanoGPTs_HMMs}, the nanoGPT models exhibit similar performance across all settings, with closely aligned validation losses, and consistently outperform the Belief Net for all $\hat{\stateDimension} \leq \stateDimension$. 
These findings indicate that even single-layer transformers can effectively capture HMM dynamics, likely due to their ability to model long-range, non-Markovian dependencies, suggesting a potential avenue for improving the Belief Net framework.

\subsection{Real-World Text Data: Character-Level Language Modeling}\label{appx:experiments_text}

In this setting (see Section~\ref{sec:real_world_text}), the Belief Net and all other models are evaluated on a language modeling task for next-token prediction using the Federalist Papers dataset.
The trained models include the Belief Net, Baum-Welch, Spectral, nanoGPT-s (single-head single-layer transformer) and nanoGPT-m (multi-head single-layer transformer).
Their performance is evaluated in terms of perplexity on the validation set, as shown in 
\figref{fig:text_results} in the main text.

The spectral method fails to identify a valid solution for this dataset under the state dimension $\stateDimension=64$, likely due to the non-Markovian nature of the text data.
A sweep over the candidate state dimension $\hat{\stateDimension}\in[1,\stateDimension]$ is performed and the model corresponding to the lowest validation loss is chosen to report the validation perplexity for the spectral method.
The final model of the spectral method has $\hat{\stateDimension}=2$ and has the highest perplexity among all methods.

\begin{figure}[t]
  \centering

  \hfill
  \begin{minipage}[b]{0.388\textwidth}
    \centering
    \includegraphics[width=\linewidth]{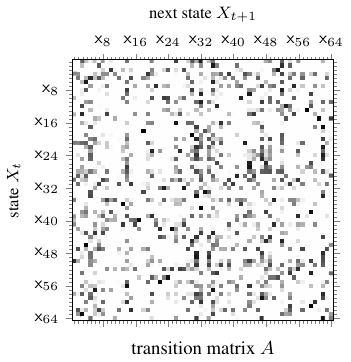}
  \end{minipage}
  \hspace{-0.9em}
  \begin{minipage}[b]{0.6\textwidth}
    \centering
    \includegraphics[width=\linewidth]{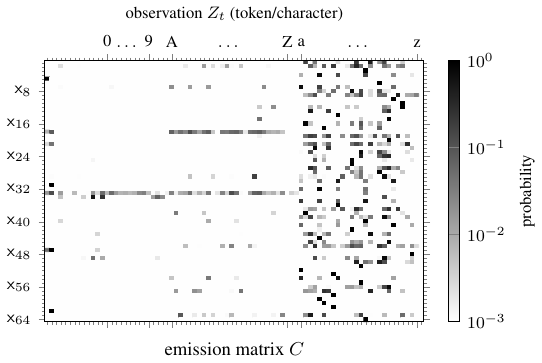}
  \end{minipage}
  \hfill%

  \caption{
    Learned transition and emission matrices of the HMM with Belief Net trained on the real-word text data: the Federalist Papers.
    The subfigure on the left is the learned transition matrix $\transitionMatrix$. 
    Each entry $\transitionMatrix_{ij}$ represents the probability of transitioning from state $\state_i$ to state $\state_j$. 
    The color intensity indicates the magnitude of the transition probabilities with darker colors representing higher probabilities. 
    The axes are labeled with the corresponding states.
    The subfigure on the right is the learned emission matrix $\emissionMatrix$. 
    Each entry $\emissionMatrix_{ik}$ represents the probability of emitting token $\observation_k$ from state $\state_i$.
    The observations $\observation_{14:23}$ correspond to the numeric characters 0-9, $\observation_{28:53}$ correspond to the uppercase letters A-Z, and $\observation_{56:81}$ correspond to the lowercase letters a-z.
    The remaining observations correspond to special characters and whitespace.
    The color intensity indicates the magnitude of the emission probabilities with darker colors representing higher probabilities. 
    The axes are labeled with the corresponding states and tokens.
    The color bars indicate the log scale of the probabilities with the minimum visualized value set to $10^{-3}$.
  }\label{fig:matrices}
\end{figure}

The learned transition and emission matrices of the HMM with Belief Net are visualized in \figref{fig:matrices}.
The transition matrix $\transitionMatrix$ shows that each state only transitions to a few other states, indicating that the model is learning a sparse transition structure.
The emission matrix $\emissionMatrix$ shows more interpretable patterns: certain states emits specific characters with high probability.
For example, state $\state_{33}$ emits the digits and uppercase letters with high probability, while state $\state_{18}$ emits uppercase letters only.
The likelihood of emitting lowercase letters is generally high across all states, which may be due to the fact that the dataset contains more lowercase letters than all other characters.

%% file: figures/hmm/undercomplete/loss.tex
\pgfmathsetmacro{\lossRandomGuess}{ln(128)}
\def\lossBaumWelsh{1.951}
\def\lossSpectral{1.624}
\def\lossNanoGPTsingle{1.458}
\def\lossNanoGPTmulti{1.475}
\def\lossFilter{1.368}

\definecolor{colorBlue}{RGB}{31,119,180}
\definecolor{colorOrange}{RGB}{255,127,14}
\definecolor{colorGreen}{RGB}{44,160,44}
\definecolor{colorRed}{RGB}{214,39,40}

\def\minIteration{0.5}
\def\maxIteration{4000}

\begin{axis}[
  width=7.5cm,
  height=5.35cm,
  ylabel={loss $\mathsf{J}_l$},
  xmode=log,
  xmin=\minIteration, xmax=\maxIteration,
  ymin=-0.6, ymax=6.6,
  xtick={1,10, 100, 1000},
  grid=major,
  grid style={dashed, gray!20},
  legend pos=north east,
  legend style={
      line width=0.3pt,
  },
  legend cell align={left},
  title={undercomplete $\observationDimension=128$},
  title style={font=\small},
  line width=1.5pt,
  tick label style={font=\footnotesize},
  label style={font=\small},
  every axis plot/.append style={thick},
  axis on top,
]


  \def\xpositionLeft{0.5}
  \def\xpositionRight{4000}

  \node[
    anchor=south west, 
    colorBlue,
    align=center,
  ] at (axis cs:4, \lossBaumWelsh) {Baum-\\Welch};
  \addplot[
    colorBlue, 
    dashed, 
    line width=1pt, 
    forget plot,
  ] coordinates {(6, \lossBaumWelsh) (\maxIteration, \lossBaumWelsh)};

  \node[
    anchor=south west, 
    colorGreen,
  ] at (axis cs:\xpositionLeft, \lossSpectral) {Spectral};
  \addplot[
    colorGreen, 
    dashed, 
    line width=1pt, 
    forget plot,
  ] coordinates {(\minIteration, \lossSpectral) (\maxIteration, \lossSpectral)};

  \node[
    anchor=north west, 
    colorOrange,
  ] at (axis cs:\xpositionLeft, \lossNanoGPTmulti) {nanoGPT};
  \addplot[
    opacity=0.7,
    colorOrange, 
    dashed, 
    line width=1pt, 
    forget plot,
  ] coordinates {(\minIteration, \lossNanoGPTmulti) (\maxIteration, \lossNanoGPTmulti)};
  \addplot[
    colorOrange, 
    dashed, 
    line width=1pt, 
    forget plot,
  ] coordinates {(\minIteration, \lossNanoGPTsingle) (\maxIteration, \lossNanoGPTsingle)};

  \node[
    anchor=south west, 
    gray,
  ] at (axis cs:\xpositionLeft, \lossRandomGuess) {random ($\ln\observationDimension$)};
  \addplot[
    gray, 
    dashed, 
    line width=1pt, 
    forget plot,
  ] coordinates {(\minIteration, \lossRandomGuess) (\maxIteration, \lossRandomGuess)};

  \node[
    anchor=north east, 
    black,
  ] at (axis cs:\xpositionRight, \lossFilter) {HMM filter (optimal)};
  \addplot[
    black, 
    dashed, 
    line width=1pt, 
    forget plot,
  ] coordinates {(6, \lossFilter) (\maxIteration, \lossFilter)};

  \addplot[
    color=colorRed,
    opacity=0.6,
    line width=1.5pt,
  ] table[
    col sep=comma,  
    x=iteration, 
    y=loss,
  ] {figures/hmm/undercomplete/training_loss.csv};

  \addplot[
    color=colorRed,
    line width=1.5pt,
  ] table[
    col sep=comma,  
    x=iteration, 
    y=loss,
  ] {figures/hmm/undercomplete/validation_loss.csv};

\end{axis}

%% file: figures/hmm/overcomplete/loss.tex
\pgfmathsetmacro{\lossRandomGuess}{ln(32)}
\def\lossBaumWelch{1.216}
\def\lossNanoGPTsingle{0.829}
\def\lossNanoGPTmulti{0.810}
\def\lossFilter{0.737}

\definecolor{colorBlue}{RGB}{31,119,180}
\definecolor{colorOrange}{RGB}{255,127,14}
\definecolor{colorGreen}{RGB}{44,160,44}
\definecolor{colorRed}{RGB}{214,39,40}

\def\minIteration{0.5}
\def\maxIteration{4000}

\begin{axis}[
  width=7.5cm,
  height=5.35cm,
  xmode=log,
  xmin=\minIteration, xmax=\maxIteration,
  ymin=-0.4, ymax=4.4,
  xtick={1,10, 100, 1000, 10000},
  grid=major,
  grid style={dashed, gray!20},
  legend pos=north east,
  legend style={
      line width=0.3pt,
  },
  legend cell align={left},
  title={overcomplete $\observationDimension=32$},
  title style={font=\small},
  line width=1.5pt,
  tick label style={font=\footnotesize},
  label style={font=\small},
  every axis plot/.append style={thick},
  axis on top,
]


  \def\xpositionLeft{0.5}
  \def\xpositionRight{4000}

  \node[
    anchor=south west, 
    colorBlue,
  ] at (axis cs:\xpositionLeft, \lossBaumWelch) {Baum-Welch};
  \addplot[
    colorBlue, 
    dashed, 
    line width=1pt, 
    forget plot,
  ] coordinates {(\minIteration, \lossBaumWelch) (\maxIteration, \lossBaumWelch)};


  \node[
    anchor=north west, 
    colorOrange,
  ] at (axis cs:\xpositionLeft, \lossNanoGPTmulti) {nanoGPT};
  \addplot[
    opacity=0.7,
    colorOrange, 
    dashed, 
    line width=1pt, 
    forget plot,
  ] coordinates {(\minIteration, \lossNanoGPTmulti) (\maxIteration, \lossNanoGPTmulti)};
  \addplot[
    colorOrange, 
    dashed, 
    line width=1pt, 
    forget plot,
  ] coordinates {(\minIteration, \lossNanoGPTsingle) (\maxIteration, \lossNanoGPTsingle)};

  \node[
    anchor=south west, 
    gray,
  ] at (axis cs:\xpositionLeft, \lossRandomGuess) {random ($\ln\observationDimension$)};
  \addplot[
    gray, 
    dashed, 
    line width=1pt, 
    forget plot,
  ] coordinates {(\minIteration, \lossRandomGuess) (\maxIteration, \lossRandomGuess)};

  \node[
    anchor=north east, 
    black,
  ] at (axis cs:\xpositionRight, \lossFilter) {HMM filter (optimal)};
  \addplot[
    black, 
    dashed, 
    line width=1pt, 
    forget plot,
  ] coordinates {(6, \lossFilter) (\maxIteration, \lossFilter)};

  \addplot[
    color=colorRed,
    opacity=0.6,
    line width=1.5pt,
  ] table[
    col sep=comma,  
    x=iteration, 
    y=loss,
  ] {figures/hmm/overcomplete/training_loss.csv};

  \addplot[
    color=colorRed,
    line width=1.5pt,
  ] table[
    col sep=comma,  
    x=iteration, 
    y=loss,
  ] {figures/hmm/overcomplete/validation_loss.csv};

\end{axis}

%% file: figures/initialization/initialization_sensitivity.tex
\hfill
\begin{minipage}{0.44\textwidth}
  \centering
  \begin{tikzpicture}[font=\footnotesize]
    \pgfplotstableread[col sep=comma]{figures/initialization/baumwelch_loss_statistics_m128.csv}\datatableBaumWelch
    \pgfplotstableread[col sep=comma]{figures/initialization/beliefnet_loss_statistics_m128.csv}\datatableBeliefNet
    \pgfmathsetmacro{\lossRandomGuess}{ln(128)}
    \def\lossFilter{1.368}
    \def\maxLoss{6.6}
    \def\minLoss{-0.6}
    \def\yLabelFlag{true}
    \def\axisTitle{undercomplete  $\observationDimension=128$}
    \input{figures/initialization/template.tex}
  \end{tikzpicture}
\end{minipage}
\begin{minipage}{0.45\textwidth}
  \centering
  \begin{tikzpicture}[font=\footnotesize]
    \pgfplotstableread[col sep=comma]{figures/initialization/baumwelch_loss_statistics_m32.csv}\datatableBaumWelch
    \pgfplotstableread[col sep=comma]{figures/initialization/beliefnet_loss_statistics_m32.csv}\datatableBeliefNet
    \pgfmathsetmacro{\lossRandomGuess}{ln(32)}
    \def\lossFilter{0.737}
    \def\maxLoss{4.4}
    \def\minLoss{-0.4}
    \def\yLabelFlag{false}
    \def\axisTitle{overcomplete $\observationDimension=32$}
    \input{figures/initialization/template.tex}
  \end{tikzpicture}
\end{minipage}

\hfill
\begin{minipage}[t]{0.88\textwidth}
  \vspace{0.25\baselineskip}
  \centering
  \small
  candidate state dimension $\hat{\stateDimension}$ ($\stateDimension=64$)
\end{minipage}

\centering
\begin{tikzpicture}[remember picture, overlay]
  \node[anchor=east, xshift=-4.2cm, yshift=0.4cm] at (0,0) {
    \begin{minipage}{4cm}
      \begin{center}
        \begin{tikzpicture}[font=\footnotesize]

          \definecolor{colorBlue}{RGB}{31,119,180}
          \definecolor{colorOrange}{RGB}{255,127,14}
          \definecolor{colorGreen}{RGB}{44,160,44}
          \definecolor{colorRed}{RGB}{214,39,40}
          \begin{axis}[
              hide axis,
              xmin=10, xmax=50, ymin=0, ymax=0.4,
              legend style={
                  draw=none,
                  fill=none,
                  at={(0.5,0.5)}, 
                  anchor=center, 
                  cells={anchor=west}, 
                  legend columns=1, 
                  /tikz/every even column/.append style={column sep=5pt},
              },
          ]

            \addlegendimage{
              color=colorBlue, 
              mark=square,
              line width=0.75pt,
            }
            \addlegendentry{\color{black!60}Baum-Welch}

            \addlegendimage{
              color=colorRed, 
              mark=diamond,
              line width=0.75pt,
            }
            \addlegendentry{\textbf{Belief Net}}

          \end{axis}
        \end{tikzpicture}
      \end{center}
    \end{minipage}
  };
\end{tikzpicture}

%% file: figures/parameter_recovery/matrix_comparison.tex
\hfill
\begin{minipage}{0.44\textwidth}
  \centering
  \begin{tikzpicture}[font=\footnotesize]
    \gdef\xLabelFlag{true} 
    \gdef\yLabelFlag{true} 
    \pgfplotstableread[col sep=comma]{figures/parameter_recovery/transition/hmm_eigs_m_128.csv}\datatableModel

    \pgfplotstableread[col sep=comma]{figures/parameter_recovery/transition/best_bw/m128_d4_eigvals.csv}\datatableBWsmall
    \pgfplotstableread[col sep=comma]{figures/parameter_recovery/transition/best_bw/m128_d16_eigvals.csv}\datatableBWmedium
    \pgfplotstableread[col sep=comma]{figures/parameter_recovery/transition/best_bw/m128_d64_eigvals.csv}\datatableBWright
    \pgfplotstableread[col sep=comma]{figures/parameter_recovery/transition/best_bw/m128_d256_eigvals.csv}\datatableBWhigh

    \pgfplotstableread[col sep=comma]{figures/parameter_recovery/transition/best_beliefnet/m128_d4_eigvals.csv}\datatableBNsmall
    \pgfplotstableread[col sep=comma]{figures/parameter_recovery/transition/best_beliefnet/m128_d16_eigvals.csv}\datatableBNmedium
    \pgfplotstableread[col sep=comma]{figures/parameter_recovery/transition/best_beliefnet/m128_d64_eigvals.csv}\datatableBNright
    \pgfplotstableread[col sep=comma]{figures/parameter_recovery/transition/best_beliefnet/m128_d256_eigvals.csv}\datatableBNhigh

    \def\axisTitle{undercomplete $\observationDimension=128$}
    \input{figures/parameter_recovery/transition/template.tex}
  \end{tikzpicture}
\end{minipage}
\begin{minipage}{0.49\textwidth}
  \centering
  \begin{tikzpicture}[font=\footnotesize]
    \gdef\xLabelFlag{true} 
    \gdef\yLabelFlag{false} 
    \pgfplotstableread[col sep=comma]{figures/parameter_recovery/transition/hmm_eigs_m_32.csv}\datatableModel

    \pgfplotstableread[col sep=comma]{figures/parameter_recovery/transition/best_bw/m32_d4_eigvals.csv}\datatableBWsmall
    \pgfplotstableread[col sep=comma]{figures/parameter_recovery/transition/best_bw/m32_d16_eigvals.csv}\datatableBWmedium
    \pgfplotstableread[col sep=comma]{figures/parameter_recovery/transition/best_bw/m32_d64_eigvals.csv}\datatableBWright
    \pgfplotstableread[col sep=comma]{figures/parameter_recovery/transition/best_bw/m32_d256_eigvals.csv}\datatableBWhigh

    \pgfplotstableread[col sep=comma]{figures/parameter_recovery/transition/best_beliefnet/m32_d4_eigvals.csv}\datatableBNsmall
    \pgfplotstableread[col sep=comma]{figures/parameter_recovery/transition/best_beliefnet/m32_d16_eigvals.csv}\datatableBNmedium
    \pgfplotstableread[col sep=comma]{figures/parameter_recovery/transition/best_beliefnet/m32_d64_eigvals.csv}\datatableBNright
    \pgfplotstableread[col sep=comma]{figures/parameter_recovery/transition/best_beliefnet/m32_d256_eigvals.csv}\datatableBNhigh

    \def\axisTitle{overcomplete $\observationDimension=32$}
    \input{figures/parameter_recovery/transition/template.tex}
  \end{tikzpicture}
\end{minipage}

\begin{minipage}{\textwidth}
  \vspace{0.25\baselineskip}
  \centering
  \small
  \hspace{8em}eigenvalue index (sorted)
\end{minipage}

\centering
\begin{tikzpicture}[remember picture, overlay]
  \node[anchor=east, xshift=-4.2cm, yshift=0.6cm] at (0,0) {
    \begin{minipage}{4cm}
      \begin{center}
        \begin{tikzpicture}[font=\footnotesize]

          \definecolor{colorBlue}{RGB}{31,119,180}
          \definecolor{colorOrange}{RGB}{255,127,14}
          \definecolor{colorGreen}{RGB}{44,160,44}
          \definecolor{colorRed}{RGB}{214,39,40}
          \begin{axis}[
              hide axis,
              xmin=10, xmax=50, ymin=0, ymax=0.4,
              legend style={
                  draw=none,
                  fill=none,
                  at={(0.5,0.5)}, 
                  anchor=center, 
                  cells={anchor=west}, 
                  legend columns=1, 
                  /tikz/every even column/.append style={column sep=5pt},
              },
          ]

            \addlegendimage{
              color=black, 
              mark=*,
              only marks,
            }
            \addlegendentry{\color{black!60}model}


            \addlegendimage{
              color=colorBlue, 
              mark=square,
              line width=0.75pt,
            }
            \addlegendentry{\color{black!60}Baum-Welch}

            \addlegendimage{
              color=colorRed, 
              mark=diamond,
              line width=0.75pt,
            }
            \addlegendentry{\textbf{Belief Net}}

          \end{axis}
        \end{tikzpicture}
      \end{center}
    \end{minipage}
  };
\end{tikzpicture}

\hfill
\begin{minipage}{0.44\textwidth}
  \centering
  \begin{tikzpicture}[font=\footnotesize]
    \gdef\xLabelFlag{true} 
    \gdef\yLabelFlag{true} 
    \pgfplotstableread[col sep=comma]{figures/parameter_recovery/emission/bw_avg_mismatch_m_128.csv}\datatableBaumWelch
    \pgfplotstableread[col sep=comma]{figures/parameter_recovery/emission/beliefnet_avg_mismatch_m_128.csv}\datatableBeliefNet
    \input{figures/parameter_recovery/emission/template.tex}
  \end{tikzpicture}
\end{minipage}
\begin{minipage}{0.49\textwidth}
  \centering
  \begin{tikzpicture}[font=\footnotesize]
    \gdef\xLabelFlag{true} 
    \gdef\yLabelFlag{false} 
    \pgfplotstableread[col sep=comma]{figures/parameter_recovery/emission/bw_avg_mismatch_m_32.csv}\datatableBaumWelch
    \pgfplotstableread[col sep=comma]{figures/parameter_recovery/emission/beliefnet_avg_mismatch_m_32.csv}\datatableBeliefNet
    \input{figures/parameter_recovery/emission/template.tex}
  \end{tikzpicture}
\end{minipage}

\begin{minipage}{\textwidth}
  \vspace{0.25\baselineskip}
  \centering
  \small
  \hspace{8em}candidate state dimension $\hat{\stateDimension}$ ($\stateDimension=64$)
\end{minipage}







%% file: figures/parameter_recovery/complete_comparison.tex
\hfill
\begin{minipage}{0.52\textwidth}
  \centering

  \begin{tikzpicture}[font=\footnotesize]
    \gdef\xLabelFlag{false} 
    \gdef\yLabelFlag{true} 
    \pgfplotstableread[col sep=comma]{figures/parameter_recovery/loss_nanogpt/m_128_lambda_07_model_m.csv}\datatableModelM 
    \pgfplotstableread[col sep=comma]{figures/parameter_recovery/loss_nanogpt/m_128_lambda_07_model_s.csv}\datatableModelS
    \pgfplotstableread[col sep=comma]{figures/parameter_recovery/loss_beliefnet/m_128_lambda_07.csv}\datatableBeliefNet
    \pgfmathsetmacro{\lossRandomGuess}{ln(128)}
    \def\lossFilter{2.062} 

    \def\maxLoss{7.2}
    \def\minLoss{-1.2}

    \def\axisTitle{undercomplete  $\observationDimension=128$}
    \def\yLabelText{$\homotopyParameter=0.7$\\(fast mixing)\\\\validation loss $\mathsf{J}$}
    \input{figures/parameter_recovery/template.tex}
  \end{tikzpicture}\\[-\baselineskip]
  \begin{tikzpicture}[font=\footnotesize]
    \gdef\xLabelFlag{true} 
    \gdef\yLabelFlag{true} 
    \pgfplotstableread[col sep=comma]{figures/parameter_recovery/loss_nanogpt/m_128_lambda_10_model_m.csv}\datatableModelM 
    \pgfplotstableread[col sep=comma]{figures/parameter_recovery/loss_nanogpt/m_128_lambda_10_model_s.csv}\datatableModelS
    \pgfplotstableread[col sep=comma]{figures/parameter_recovery/loss_beliefnet/m_128_lambda_10.csv}\datatableBeliefNet
    \pgfmathsetmacro{\lossRandomGuess}{ln(128)}
    \def\lossFilter{0.854} 

    \def\maxLoss{7.2}
    \def\minLoss{-1.2}

    \def\axisTitle{}
    \def\yLabelText{$\homotopyParameter=1.0$\\(slow mixing)\\\\validation loss $\mathsf{J}$}
    \input{figures/parameter_recovery/template.tex}
  \end{tikzpicture}

\end{minipage}
\begin{minipage}{0.45\textwidth}
  \centering

  \begin{tikzpicture}[font=\footnotesize]
    \gdef\xLabelFlag{false} 
    \gdef\yLabelFlag{false} 
    \pgfplotstableread[col sep=comma]{figures/parameter_recovery/loss_nanogpt/m_32_lambda_07_model_m.csv}\datatableModelM 
    \pgfplotstableread[col sep=comma]{figures/parameter_recovery/loss_nanogpt/m_32_lambda_07_model_s.csv}\datatableModelS
    \pgfplotstableread[col sep=comma]{figures/parameter_recovery/loss_beliefnet/m_32_lambda_07.csv}\datatableBeliefNet
    \pgfmathsetmacro{\lossRandomGuess}{ln(32)}
    \def\lossFilter{1.448} 

    \def\maxLoss{4.8}
    \def\minLoss{-0.8}

    \def\axisTitle{overcomplete  $\observationDimension=32$}
    \def\yLabelText{}
    \input{figures/parameter_recovery/template.tex}
  \end{tikzpicture}\\[-\baselineskip]
  \begin{tikzpicture}[font=\footnotesize]
    \gdef\xLabelFlag{true} 
    \gdef\yLabelFlag{false} 
    \pgfplotstableread[col sep=comma]{figures/parameter_recovery/loss_nanogpt/m_32_lambda_10_model_m.csv}\datatableModelM 
    \pgfplotstableread[col sep=comma]{figures/parameter_recovery/loss_nanogpt/m_32_lambda_10_model_s.csv}\datatableModelS
    \pgfplotstableread[col sep=comma]{figures/parameter_recovery/loss_beliefnet/m_32_lambda_10.csv}\datatableBeliefNet
    \pgfmathsetmacro{\lossRandomGuess}{ln(32)}
    \def\lossFilter{0.222} 

    \def\maxLoss{4.8}
    \def\minLoss{-0.8}

    \def\axisTitle{}
    \def\yLabelText{}
    \input{figures/parameter_recovery/template.tex}
  \end{tikzpicture}
\end{minipage}

\begin{minipage}[t]{\textwidth}
  \vspace{0.25\baselineskip}
  \centering
  \small
  \hspace{7em}candidate state dimension $\hat{\stateDimension}$ ($\stateDimension=64$)
\end{minipage}

\begin{tikzpicture}[remember picture, overlay]
  \node[anchor=east, xshift=-4.2cm, yshift=0.7cm] at (0,0) {
    \begin{minipage}{4cm}
      \begin{center}
        \begin{tikzpicture}[font=\footnotesize]

          \definecolor{colorBlue}{RGB}{31,119,180}
          \definecolor{colorOrange}{RGB}{255,127,14}
          \definecolor{colorGreen}{RGB}{44,160,44}
          \definecolor{colorRed}{RGB}{214,39,40}
          \begin{axis}[
              hide axis,
              xmin=10, xmax=50, ymin=0, ymax=0.4,
              legend style={
                  draw=none,
                  fill=none,
                  at={(0.5,0.5)}, 
                  anchor=center, 
                  cells={anchor=west}, 
              },
          ]

            \addlegendimage{
              color=colorOrange, 
              mark=triangle,
              line width=0.75pt,
            }
            \addlegendentry{\color{black!60}nanoGPT-s}

            \addlegendimage{
              opacity=0.7,
              color=colorOrange, 
              mark=triangle,
              line width=0.75pt,
            }
            \addlegendentry{\color{black!60}nanoGPT-m}

            \addlegendimage{
              color=colorRed, 
              mark=diamond,
              line width=0.75pt,
            }
            \addlegendentry{\textbf{Belief Net}}

          \end{axis}
        \end{tikzpicture}
      \end{center}
    \end{minipage}
  };
\end{tikzpicture}